\documentclass[11pt,a4paper,oneside,twocolumn]{article}

\usepackage{cvpr-geom}
\usepackage{cvpr-abbriv}

\usepackage[small,compact]{titlesec}

\usepackage[rreport]{cmpcover}

\usepackage[pagebackref=true,breaklinks=true,letterpaper=true,colorlinks,bookmarks=false]{hyperref}

\usepackage{alg}
\usepackage{epsfig}
\usepackage{graphicx}
\usepackage{amsmath}
\usepackage{amssymb}

\usepackage{amsthm}
\usepackage{amsfonts}
\usepackage{color}
\usepackage{epsfig}

\newcommand{\<}{\langle}
\renewcommand{\>}{\rangle}

\newcommand{\Real}{\mathbb{R}}

\newcommand{\tab}{{\phantom{bla}}}

\newcommand{\V}{\mathcal{V}}
\newcommand{\U}{\mathcal{U}}

\newcommand{\E}{\mathcal{E}}

\renewcommand{\L}{\mathcal{L}}

\newcommand{\st}{\mbox{s.t. }}

\newcounter{myRomanCounter}

\newcommand{\bzero}{{\bf 0}}
\newcommand{\x}{{\bf x}}
\newcommand{\y}{{\bf y}}

\newcommand{\w}{{\bf w}}
\newcommand{\bc}{{\bf c}}

\theoremstyle{plain}

\theoremstyle{definition}

\newcommand{\AS}{Alexander Shekhovtsov}

\newcommand{\red}{\color[rgb]{1,0,0}}
\newcommand{\green}{\color[rgb]{0,1,0}}

\newlength{\figwidth}
\newlength{\figwidtha}

\newcommand*{\mypar}[1]{\subsection{#1}}
\newcommand{\sectionskip}{\vspace{-1mm}}
\title{Curvature Prior for MRF-based Segmentation and Shape Inpainting}

\author{\AS{\small$^1$} \and Pushmeet Kohli{\small$^2$}  \and Carsten Rother{\small$^2$}}% \and \VH{\small$^1$}}
\CMPEmail{shekhole@fel.cvut.cz \and pkohli@microsoft.com \and carrot@microsoft.com} %\and hlavac@fel.cvut.cz}
\CMPAffiliation{
{\small$^1$}Center for Machine Perception, K13133 FEE Czech Technical
                  University\and
{\small$^2$}Microsoft Research Cambridge
}

\CMPReportNo{CTU--CMP--2011--11}
\CMPAcknowledgement{A. Shekhovtsov was supported by EU project FP7-ICT-247870 NIFTi.}
\defRCSRevision{$Revision: 1.0 $}
\CMPSubtitle{(Version \RCSRevision)}
\begin{document}

\begin{abstract}
Most image labeling problems such as segmentation and image reconstruction are fundamentally ill-posed and suffer from ambiguities and noise. Higher order image priors encode high level structural dependencies between pixels and are key to overcoming these problems. However, these priors in general lead to computationally intractable models. This paper addresses the problem of discovering compact representations of higher order priors which allow efficient inference.
We propose a framework for solving this problem which uses a recently proposed representation of higher order functions where they are encoded as lower envelopes of linear functions. Maximum a Posterior inference on our learned models reduces to minimizing a pairwise function of discrete variables, which can be done approximately using standard methods. Although this is a primarily theoretical paper, we also demonstrate the practical effectiveness of our framework on the problem of learning a shape prior for image segmentation and reconstruction. We show that our framework can learn a compact representation that approximates a prior that encourages low curvature shapes. We evaluate the approximation accuracy, discuss properties of the trained model, and show various results for shape inpainting and image segmentation.

\end{abstract}
%%%%%%%%% BODY TEXT
%\vspace{-3mm}
\section{Introduction}\label{introduction}
Most computer vision problems can be formulated in terms of estimating the values of hidden variables from a given set of observations. In such a setting, probabilistic models are applied to represent the prior knowledge about hidden variables and their statistical relationship with observed variables.

\begin{figure}[tr]
\begin{center}
\includegraphics[width=\columnwidth]{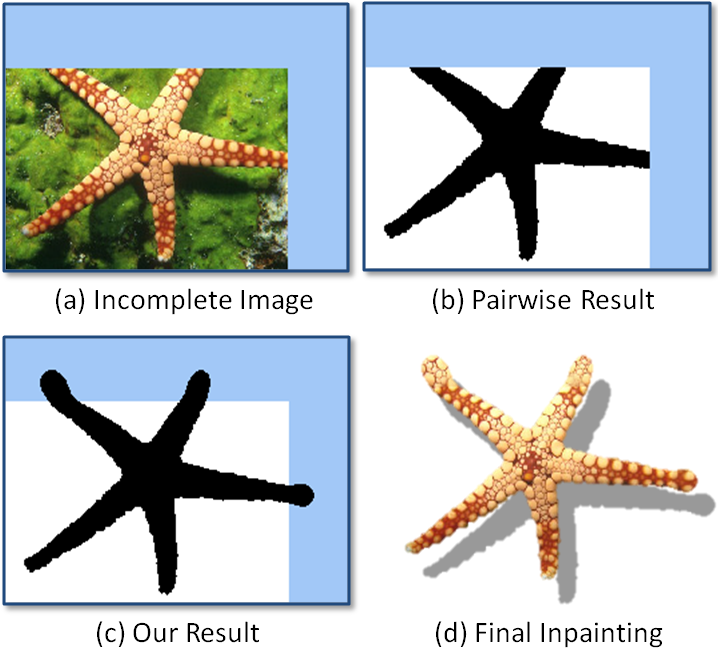}
\caption{\it (a) Input image (area for completion of starfish is shown in blue). (b) The starfish was interactively segmented from the image. Then the three arms  of the starfish, which touch the image borders, were completed with an 8-connected pairwise MRF which encodes a standard length prior. Note, with this prior no pixels in the blue completion area were assigned to the starfish arms. (c) Completion of the shape of the three starfish arms was done with our compact-higher-order prior which models curvature. (d) Finally, texture was added fully automatically using \cite{BarnesSFG09}.}
\label{fig:teaser}
\end{center}
\vspace{-4mm}
\end{figure}

A number of models encoding prior knowledge about scenes have been proposed in computer vision. The most popular ones have been in the form of a pairwise Markov Random Field (MRF). A {\em random field} is a strictly positive probability distribution of a collection of random variables. {\em Markov Random Field} (MRF) additionally satisfies some (or none) Markov (conditional independence) properties~\cite{Lauritzen96}. An important characteristic of an MRF is the factorization of the distribution into a product of factors. {\em Pairwise} MRFs can be written as a product of factors defined over two variables at a time.  For discrete variables, this enables non-parametric representation of factors and the use of efficient optimization algorithms for approximate inference of the Maximum-a-Posteriori (MAP) solution. However, because of their restricted pairwise form, the model is not able to encode many types of powerful structural properties of images. Curvature is one such property which is known to be extremely helpful for inpainting (see figure \ref{fig:teaser}), segmentation, and many other related problems.

\begin{figure*}[tr]
\begin{center}
\setlength{\figwidth}{0.58\linewidth}
\setlength{\tabcolsep}{3pt}
\setlength{\doublerulesep}{0pt}
%\input{}
%\fbox{
\begin{tabular}{ccc}
\begin{tabular}{c}\hspace{-1mm}\includegraphics[width=0.42\figwidth]{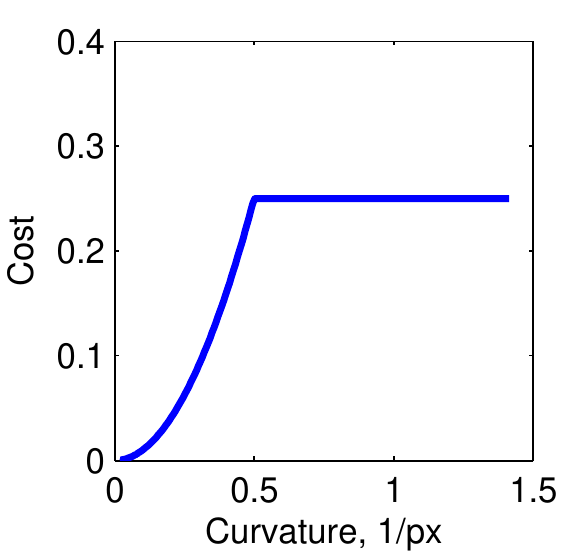}\end{tabular}&
\begin{tabular}{c}\includegraphics[width=0.7\figwidth]{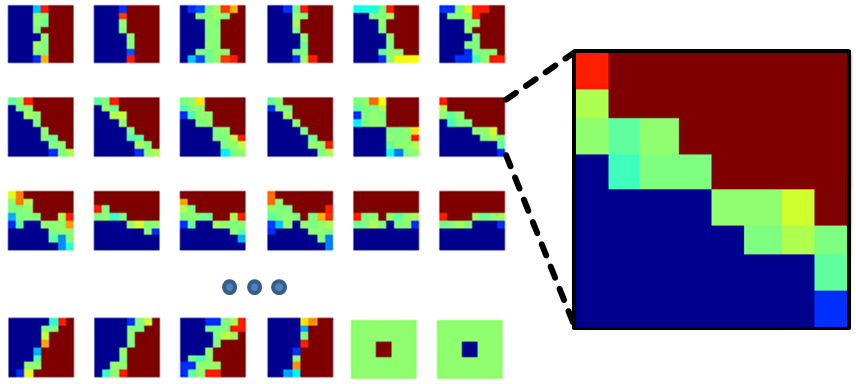}\end{tabular} &
\begin{tabular}{c}\includegraphics[width=0.44\figwidth]{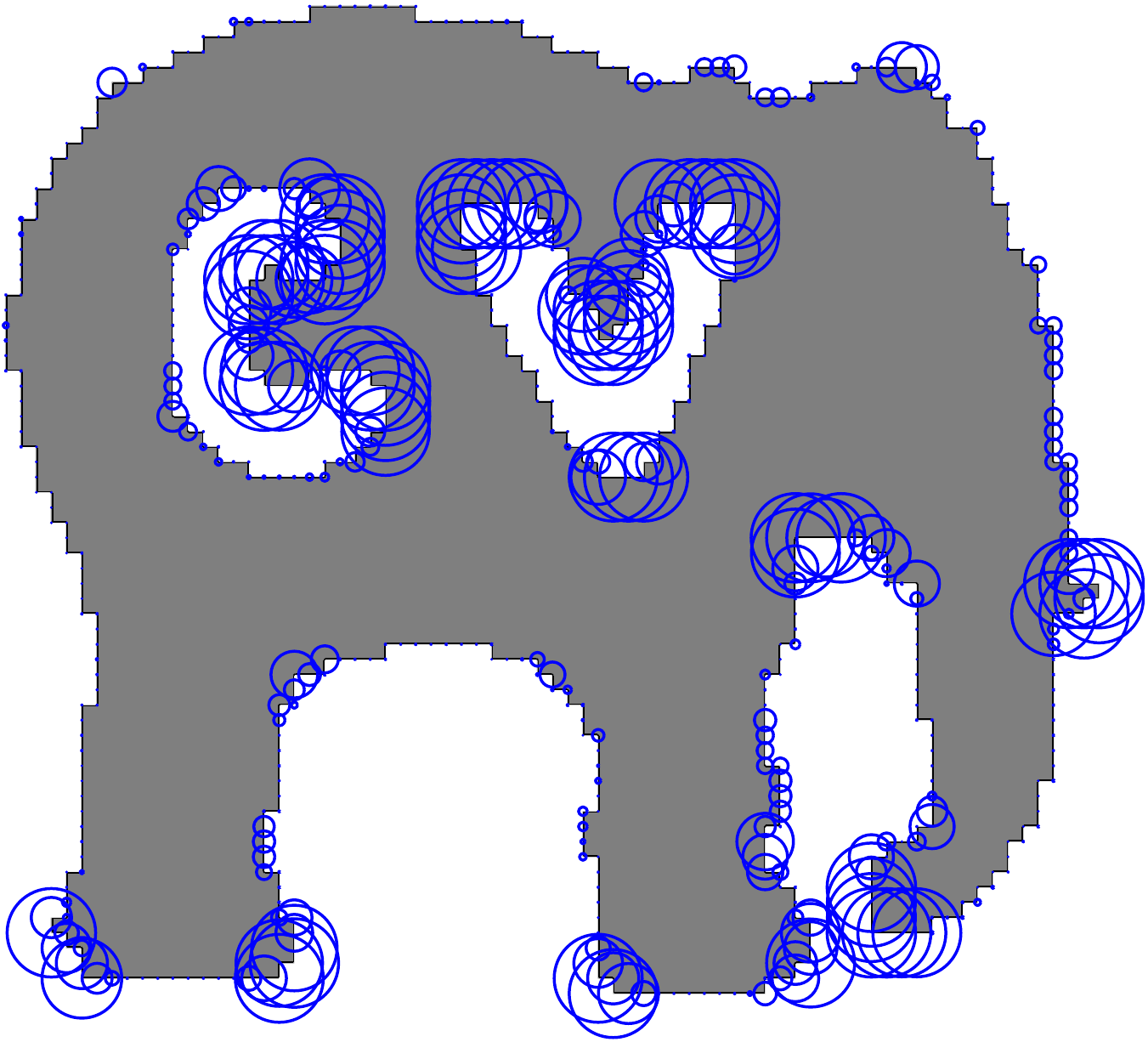}\end{tabular}\\
(a) & (b) & (c) \vspace{-2mm}
\end{tabular}
\center
%\begin{tabular}{c}\includegraphics[width=0.5\linewidth]{fig/dummy}\end{tabular}%
%\end{tabular}
%\includegraphics[width = \linewidth, height = \linewidth]{fig/dummy}
%}
\caption{ \it
(a) A given cost function for curvature, which we want to approximate.
(b) Our method learns a large set of `soft' pattern-based potentials which
implicitly model the curvature cost. Our learned MRF model has a higher-order
$8 \times 8$ pattern-based potential at each pixel-location.
%The higher order
%potentials are of the form $\min_{y}[\w_y \x_V +c_j]$ and operate  directly on the
%pixel labeling $\x_V$ corresponding to local window $V$. Linear coefficients $\w_y$
%of the potentials are visualized in (b) and called patterns.
Our model implicitly selects at each position the best-fitting pattern for a labeling. Intuitively, a pattern fits well (has low cost) if all foreground pixels match to blue pattern weights
and all background pixels match to red pattern weights. Pixels which match to green pattern weights do not contribute and may be either background or foreground. The last two patterns encode the fact that if the higher-order potential is defined on top of a non-boundary location (all 4 center pixels are foreground or background), then the curvature cost is 0, i.e. is ignored.
(c) An example demonstrating the curvature cost computed by our pattern based approximation at different parts of the object boundary. Circle radius correspond to the assigned cost.
}
\label{Fig1}
\vspace{-8mm}
\end{center}
\end{figure*}

\paragraph{Higher-order Priors}
There has been a lot of research into priors based on high-level structural dependencies between pixels such as curvature. These priors can be represented in the probabilistic model using factors which may depend on more than two variables at a time. The largest number of variables in a factor is called the {\em order} of the probabilistic model. Higher-order factors defined on discrete variables are computationally expensive to represent. In fact, the memory and time complexity for inferring the MAP solution with general inference algorithms grows exponentially with the order, and thus has limited the use of such models. The situation is a bit different for parametric models with continuous variables. Higher-order prior models such as Product of Experts (PoE)~\cite{Hinton-99} or Field of Experts (FoE)~\cite{Roth-09-FoE} are differentiable in both parameters and hidden variables. These models thus enable inference using local gradient descent, and have led to impressive results for problems such as image restoration and optical flow.

Recent research on discrete higher-order models has focused on identifying families of higher-order factors which allow efficient inference. The factors can be categorized into 3 broad categories: (a) {\em Reducable factors}, which allow MAP inference to be reduced to the problem of minimizing a pairwise energy function of discrete variables with the addition of some auxiliary variables~\cite{KohliKumar_2010,Kohli-Ladicky-Torr-09,Kolmogorov-what-energies,Ramalingam-08,Rother09}, (b) {\em Message-enabled factors}, which allow efficient message computation and thus allow inference using message passing methods such as Belief Propagation (BP) and Tree Reweighted message passing (TRW)~\cite{Gupta07,Komodakis09,Tarlow08}, and (c) {\em Constraint factors}, which impose global constraints that can be imposed efficiently in a relaxation framework~\cite{Nowozin09,Vince09}.

\paragraph{Pattern-based Representation}
Pattern and lower-envelope based representations proposed in~\cite{KohliKumar_2010,Komodakis09,Rother09} can represent some families of {\em Reducable factors}. The higher-order potentials of~\cite{Komodakis09,Rother09} are defined by enumerating important configurations (patterns) in a local window. The model of~\cite{Rother09} additionally enables deviations from encoded patterns, by using linear weighting functions. The above models are generalized by the representation proposed in \cite{KohliKumar_2010} which encodes higher-order functions as lower (or upper) envelopes of linear (modular) functions of the variables. The complexity of representing and performing inference depends on the number of linear functions (or patterns) used for representing the higher-order factor. A number of higher-order priors can be encoded using few linear functions (or patterns) and thus allow efficient inference. However, the use of a general higher-order prior would require exponential (in the order of the factor) number of linear functions (or patterns).

\paragraph{Our Contribution}
This paper addresses the problem of discovering a compact representation of a general higher-order factor. More specifically, given a particular higher-order factor, we try to find the linear-envelope representation which best approximates it. Given a set of training examples of labeling and their corresponding desired costs, we find parameters of a linear-envelope representation that matches these costs. While the problem is difficult, we propose a simple yet effective algorithm.

We demonstrate the efficacy of our method on the problem of finding a compact `curvature prior' for object boundaries. This prior encourages smooth boundaries by assigning a high cost to high curvature shape and a low cost of low curvature shapes. Given a set of training shapes, we find parameters of a linear-envelope representation that closely match their pre-specified curvature based cost\footnote{The purpose of this exercise is only to demonstrate the power of our representation scheme. A more difficult problem would be to learn the linear envelope model in an unsupervised way from examples of segmentations. We do not address this problem in this paper.}. Figure~\ref{Fig1} illustrates our discovered model. We then use the discovered higher-order factors in the problems of object segmentation and completion. The experimental results demonstrate that incorporation of these priors leads to much better results than those obtained using low-order (pairwise MRF) based models (see figure \ref{fig:teaser}) and other state-of-the-art curvature formulations.

An outline of the paper follows. In section 2, we provide the notation, define the higher-order model, and explain the lower-envelope representation of higher-order factors. Section 3 reviews research on using curvature priors for labeling problems. Section 4 explains how we learn a lower-envelop representation of a curvature based higher-order prior model. Section 5 discusses the techniques we used to perform MAP inference in the pairwise model corresponding to the discovered higher-order model. Section 6 describes our experimental setup and provides the results. We conclude by summarizing our framework and listing some directions for future work in Section 7. 
\sectionskip
\section{Higher-order Model Representation}\label{model}
\begin{figure}[ht]
\begin{center}
\setlength{\figwidth}{0.35\linewidth}
\begin{tabular}{ccc}
(a)\ \ &
\begin{tabular}{c}
\includegraphics[width=\figwidth]{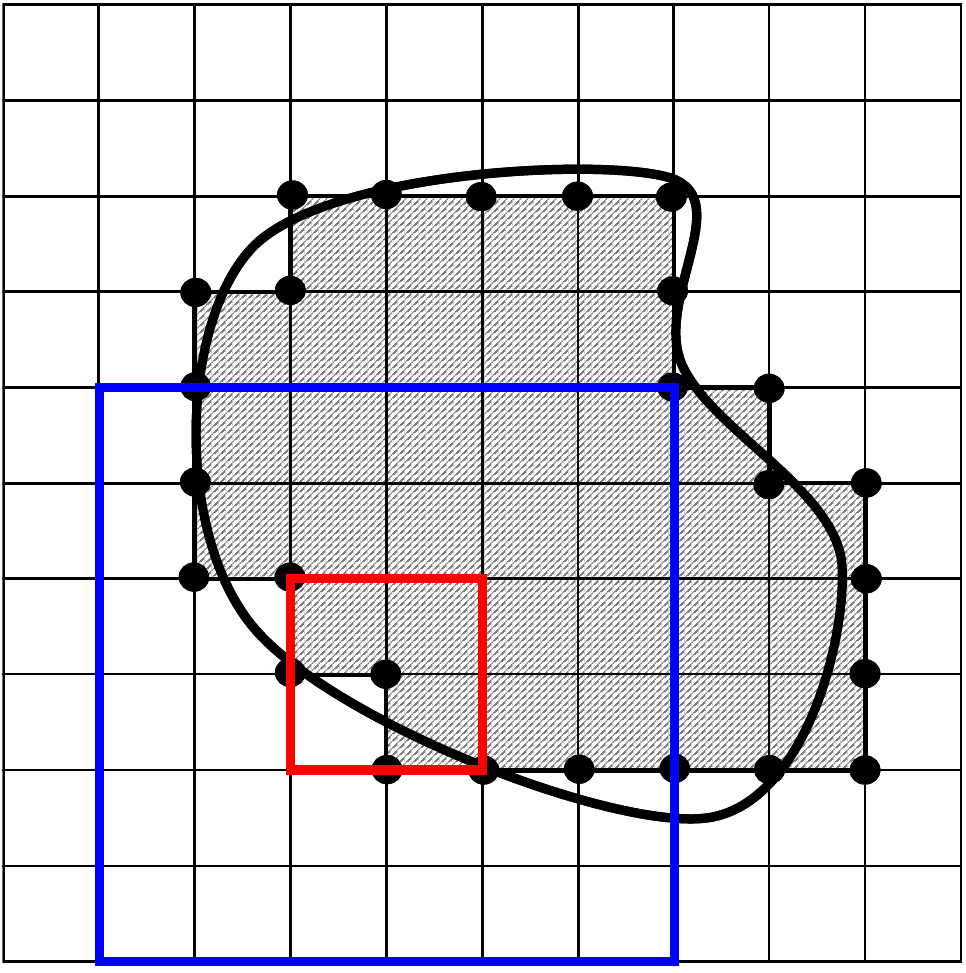}\\
\end{tabular}
&\ \ \
\begin{tabular}{cc}
\begin{tabular}{c}\includegraphics[width = 0.5\figwidth]{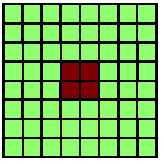}\end{tabular}&
(b)\\
\begin{tabular}{c}\includegraphics[width = 0.5\figwidth]{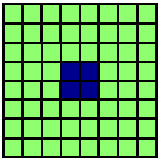}\end{tabular}&
(c)\\
\end{tabular}
\end{tabular}
\end{center}
\caption{(a) Continuous shape and its discretization. Filled circles show boundary locations. Larger blue window illustrates $V(h)$ at location $h$. (b,c) Foreground and background patterns: green $w_{y,v} = 0$, red: $w_{y,v} = +B$, blue: $w_{y,v} = -B$, constants $c_y$ are $-4B$ and $+4B$ respectively.%For convenience, we visualize locations as the centers of the even-sized windows.
%\vspace{-0.4cm}
}
\label{fig_bnd_shema}
%\end{wrapfigure}
\end{figure}

\sectionskip
We consider a set of pixels $\V=\{1\dots N_{X}\}\times \{1\dots N_{Y}\}$ and a binary set of labels $\L=\{0,1\}$, where $1$ means that a pixel belongs to the foreground (shape) and $0$ to the background. Let $\x\colon \V\to \L$ be the labeling for all pixels with individual components denoted by $x_v$, $v\in\V$. Furthermore, let $V(h) \subset \V$ denote a square window of size $K \times K$ at location $h$, and $\U$ is the set of all window locations. Windows are located densely in all pixels. More precisely, all possible $K \times K$ windows are considered which are fully inside the 2D-grid $\V$ (fig.~\ref{fig_bnd_shema} illustrates boundary locations). Let $\x_{V(h)}\colon V(h)\to \L$ denote a restriction of labeling $\x$ to the subset $V(h)$.
\par
We consider distribution of the form $p(\x) \propto \exp -E(\x)$ with the following energy function:
\begin{equation}\label{energy}
E(\x) = \sum_{v \in \V} \theta_v(x_v) + \sum_{uv \in \E}\theta_{uv}(x_u,x_v) + \sum_{h\in \U}E_h(\x),
\end{equation}
where notation $uv$ stands for ordered pair $(u,v)$, $\theta_v\colon \L\to \Real$ and $\theta_{uv}\colon \L^2\to \Real$ are unary and pairwise terms, $\E\subset \V\times\V$ is a set of pairwise terms and $E_h$ are higher-order terms. We consider the higher order terms $E_h$ of the following form (equivalent to~\cite{Rother09})
\begin{equation}\label{HOterm}
E_h(\x) = \min_{y\in P}\Big( \<\w_{y}, \x_{V(h)}\> + c_{y} \Big).
\end{equation}
This term is the minimum (lower envelope) of several modular functions of $\x_{V(h)}$. We call this model a {\em maxture} by the analogy with the mixture model discussed below. %This model indeed has some similarities with a mixture model where energy terms are log-sum-exp of linear functions (see~\cite{TR} for details).
We refer to individual linear functions $\<\w_{y}, \x_{V(h)}\> + c_{y}$ as ``soft'' patterns\footnote{The name ``soft'' refers to the fact that weights $w_{y}$ can take arbitrary values. This is in contrast to other models, ~\cite{Rother09,Komodakis09}, which constrain the weights $w_{y}$ to certain values, as discussed later.}. Here $\w_{y}\in\Real^{K^2}$ is a weight vector and $c_{y}\in \Real$ is a constant term for the pattern. Vector $\w_{y}$ is of the same size as the labeling patch $\x_{V(h)}$ and it can be visualized as an image (see fig.~\ref{Fig1}(b)). The variable $y \in P$ is called a pattern-switching variable. It is a discrete variable from the set $P=\{0,...,N_{P}\}$. We let the pattern which corresponds to $y=0$ have the associated weights $\w_{0} = \bzero$. This pattern assigns a constant value $c_{0}$ to all labelings $\x_{V(h)}$ and it ensures that $E_h(\x) \leq c_{0}$ for all $\x$. It will be needed to express models~\cite{Rother09,Komodakis09} in the form of~\eqref{HOterm}, as these models explicitly define a cut-off value. It is also used in our curvature model, where it represents the maximal cost $f^{\rm max}$ of the curvature cost function.
\par
The minimization problem of energy~\eqref{energy} expresses as
\begin{equation}
\min_\x \Big[ E_{0}(\x) + \sum_{h\in U} \min_{y\in P} \Big( \<\w_y, \x_{V(h)}\> + c_{y} \Big) \Big] , \\
\end{equation}
where unary and pairwise terms are collected into $E_0$.
The problem can also be written as a minimization of a pairwise energy
\begin{equation}\label{energy-pairwise}
\min\limits_{\substack{\x\in \L^\V \\[1pt] \y\in P^\U}} \big[ E_{0}(\x) + \sum_{h}\<\w_{y_h},\x_{V(h)}\> + c_{y_h} \big],
\end{equation}
where $\y: \U\to P$ is the concatenated vector of all pattern switching variables\footnote{We refer to components of $\y$ by $y_h$, while $y$ usually denotes an independent bound variable.}. Clearly, problem~\eqref{energy-pairwise} is a minimization of a pairwise energy function of discrete variables
$\x,\y$.
\par
Problem~\eqref{energy-pairwise} is NP-hard in general. A subclass~\cite{Kohli-Ladicky-Torr-09} where minimization of~\eqref{energy} is solvable in polynomial time is described in Appendix~\ref{subclass}. However, this class is very restrictive and is not suitable for our purpose. In the general case a number of approximate MAP inference techniques for pairwise energies can be used, as discussed in sec. \ref{optimization}.
\par
%A natural question, regarding~\eqref{energy-pairwise} is that solving this problem can be interpreted as a MAP inference of $\x$ and $\y$ jointly in a {\em pairwise} MRF in $\x$ and $\y$. However, the model for $\x$ would then be implied $\sum_y$
%This interpretation is possible but undesirable as $y$ is not a random variable here and we are not interested in estimating $y$.
%
%
%
\mypar{Pattern-based model} Let us relate the above model to the ``hard'' pattern-based model defined in~\cite{Komodakis09} (and also special case in~\cite{Rother09}). In~\cite{Komodakis09} a potential of the following form is used:
\begin{equation}
E_h(\x) =
\begin{cases}\label{hard_patterns}
\tilde c_{y}\tab \mbox{if}\ \exists y\in\{1\dots N_P\} \ \  \x_{V(h)} = {\bf p}^{y}\\
\tilde c_0\tab \mbox{otherwise}
\end{cases}.
\end{equation}
This potential assigns cost $\tilde c_{y}$ if the labeling matches exactly pattern ${\bf p}^{y} \in \L^{K^2}$ for some $y\in\{1\dots N_P\}$ and cost $\tilde c_0$ if none of the patterns are matched. The set of labels for this model is not necessarily binary. For binary labels~\eqref{hard_patterns} can be rewritten in the form~\eqref{HOterm} by setting
\begin{equation}
\begin{array}{rl}
w_{y,v} =& \begin{cases}
-B,\tab p_v^{y} = 1\\
+B,\tab p_v^{y} = 0
\end{cases}\\
c_{y} = &\tilde c_{y} + B\sum_v p_v^y\\
w_{0,v} =& 0\\
c_0 =& \tilde c_0 \ ,
\end{array}
\end{equation}
where $B$ is a sufficiently large constant. This is a restricted model since deviations from the ``hard'' patterns are not allowed, in contrast to our model~\eqref{HOterm}.
However, this restricted model allows for an alternative optimization approach which was proposed in \cite{Komodakis09} and seems to correspond to a tighter relaxation than the standard relaxation for the pairwise model~\eqref{energy-pairwise}. Also, the hard pattern potential model~\cite{Komodakis09} is too restrictive in the following sense.
Function~\eqref{HOterm}, as well as the special case~\eqref{hard_patterns}, can exactly represent an arbitrary function of discrete variables $\x_{V(h)}$ if we allow $2^{|V(h)|}$ patterns. Obviously, in practice if $|V(h)|$ is large such an approach becomes computationally infeasible. The challenge is therefore to define a good model with a small number of patterns, for which case model~\eqref{HOterm} seems clearly a better choice.
\mypar{Relation to Mixture Model}\label{model_mix}
Minimization of a pairwise energy with auxiliary variables~\eqref{energy-pairwise} can be interpreted as MAP inference in a pairwise MRF. The question arise then: why do we talk about higher order terms, rather than just introducing more hidden variables in a pairwise model? To answer this question we need to look at the associated probability distributions and the estimation problems. Here, for simplicity, we ignore unary and pairwise terms of $E_0$ and also ignore the observable variables. Let $\y$ be auxiliary hidden variables in the following, new pairwise model 
\begin{equation}
p(\x,\y) \propto \prod_{h\in U} \exp(-\<\w_{y_h},\x_{V(h)}\>-c_{y_h} ) \ .
\end{equation}
The model of $\x$ is then implied to be
\begin{equation}\label{p-mixture}
\begin{split}
p^{\rm mix}(\x) \propto \sum_{\y\in P^U} \prod_{h\in U} \exp(-\<\w_{y_h},\x_{V(h)}\>-c_{y_h})\\
= \prod_{h\in U} \sum_{y\in P} \exp(-\<\w_y,\x_{V(h)}\>-c_y).
\end{split}
\end{equation}
It can be seen that factors in this model are {\em mixtures} of exponential distributions.
The problem of MAP inference of $\x$, taking the logarithm, can be written as
\begin{equation}\label{MAP-mixture}
\arg \max_{\x} \sum_{h} \bigoplus_{y\in P}{}^{1}(-\<\w_y,\x_{V(h)}\>-c_y),
\end{equation}
where we define {\em log-sum-exp} operation $\oplus{}^\beta$ as
\begin{equation}
a\oplus^\beta b = \frac{1}{\beta} \log(e^{\beta a}+e^{\beta b}) , 
\end{equation}
which is commutative and associative binary operation, so $\bigoplus\limits_{y\in P}{}^\beta a_y$ is also unambiguously defined. The problem~\eqref{MAP-mixture} is a discrete optimization with difficult objective, so by introducing auxiliary hidden variables we arrived at a complicated inference problem. A heuristic could be used to infer $\x$ by solving the joint MAP in $\x$ and $\y$ and then to discard the estimate of $\y$, this would lead to the optimization problem of the desired form:
\begin{equation}\label{MAP-MAP-mixture}
\arg \min_{\x,\y} \sum_{h} (\<\w_{y_h},\x_{V(h)}\>+c_{y_h}).
\end{equation}
So we could have learned a mixture model and made a wrong use of it by replacing inference with~\eqref{MAP-MAP-mixture}. We prefer instead to state the model as
\begin{equation}
p^{\rm max}(\x) = \prod_{h} \max_{y\in P}\exp(-\<\w_y,\x_{V(h)}\>-c_y),
\end{equation}
where the factors are {\em maxtures} of exponential distributions. Clearly, the model corresponds to~\eqref{energy-pairwise} by rewriting it as
\begin{equation}
 = \exp\Big\{ -\sum_{h} \min_{y\in P}(\<\w_y,\x_{V(h)}\>+c_y) \Big\}.
\end{equation}
The MAP inference of $\x$ in this model directly correspond to~\eqref{MAP-MAP-mixture}. So this is a much cleaner correspondence of the model to the estimation problem. In fact, there is a smooth transition between the two models. It is know that $\lim_{\beta\to \infty} a\oplus{}^\beta b = \max(a,b)$, which says that as distributions get sharper, their mixture turns into a maximum. It is also easy to see that under this limit the model $p^{\rm mix}$ transits to $p^{\rm max}$ and so is the corresponding MAP $\x$ problem.

\section{Curvature priors for Image Labeling} \label{related}
We will evaluate the usefulness of our curvature prior on the closely related problems of image segmentation and shape inpainting. %\footnote{Shape inpainting is a practically important task since objects are very often occluded (fig.~\ref{fig_inpaint_real}), self-occluded or not full visible (fig.~ref{fig:teaser}). In the context of cut-and-pasting objects across images occluded parts have to be recovered.}.
Given an image region with a lack of observations, a good segmentation model should complete the segmentation in this region from the evidence outside of the region. This {\em shape inpainting} problem is related to inpainting of binary images which has been approached in the continuous setting with several curvature-related functionals~\cite{Bertozzi07,Chan01-CCD}\footnote{There is a vast literature on the general image inpainting problem, however these techniques, especially exemplar-based ones, do not extend to image segmentation problem, and are not relevant in the context of this paper.}.

Image labeling with curvature regularization is an important topic of research, and both continuous and discrete formulations for the problem have been proposed. Continuous formulations offer accurate models, but they rely on numerical schemes which have to deal with highly nonlinear functions and need a good initialization to converge to a good local minimum of the cost functional\footnote{For instance,~\cite{Esedoglu03-cont-segment-curv} works with discretized Euler-Lagrange equations of the 4th order.}. Discrete methods for image labeling with curvature regularization build on quantization and enumeration of boundary elements. Until recently, they were applied only in restricted scenarios where it is possible to reduce the problem to a search of the minimal path or minimum ratio cycle~\cite{Schoenemann-Cremers-07b}. These cases enjoy global optimality, however they restrict the topology or do not allow for arbitrary regional terms.

%
%\begin{wrapfigure}{r}{0.5\linewidth}
Recently,~\cite{Schoenemann09} proposed a discrete method for a general setting. This method is able to find globally optimal solutions for difficult segmentation problems. They formulate the problem as Integer Linear Programming (ILP), where variables are indicators of edge and region elements, while constraints make sure that these variables are consistent and do correspond to a shape. However, this method quantizes the directions of boundary elements, which, as we show in the experiment section, may result in large errors in the final segmentation. The complexity of the model~\cite{Schoenemann09} grows very fast with the number of directions, and it is not entirely clear how to build a cell complex with the required properties for more directions than~\cite{Schoenemann09} considers.
\par
A recent work~\cite{El-Zehiry-10} claims to give fast optimal solution for curvature regularization. However, their model is a crude approximation to the curvature functional. Its 4-neighborhood variant essentially penalizes the number of ``corners'' in the segmentation  -- locations where a $2{\times}2$ window has 3 pixels foreground and 1 background or vice-versa. It assigns zero penalty for horizontal and vertical boundaries but diagonal boundaries have maximal penalty. The 8-neighborhood variant allows for diagonal lines at zero cost but penalizes vertical and horizontal lines. Our model with $2{\times}2$ windows can also implement the 4-neighborhood model~\cite{El-Zehiry-10}. However, as we argue, a larger window is necessary to capture the curvature of a shape represented by binary pixel labeling.

\sectionskip
\section{Learning a Curvature Cost Model}\label{learning}
\sectionskip

\begin{figure}[tr]
\begin{center}
\includegraphics[width=0.9\columnwidth]{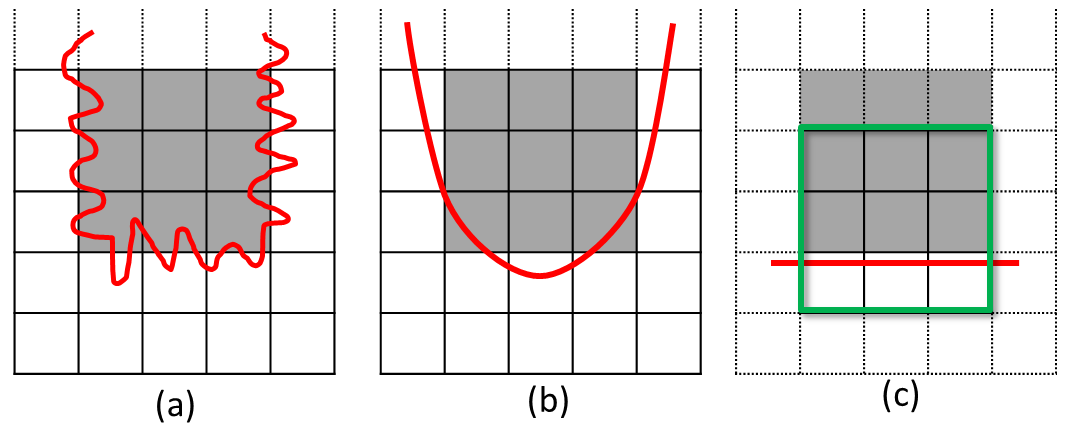}
\caption{\it {\bf Problem definition and motivation of large-sized windows.}  Examples above show discrete labelings on a pixel grid with a corresponding red continuous curve. Note, there are infinitely many continuous curves which give rise to the same discrete labelling - two examples are given in (a) and (b). The red curve in (b) is probably the one with lowest curvature given the discrete labelling. Our goal is to find an energy function which maps every discrete labelling to the corresponding cost of the continuous curvature with {\it lowest} curvature.  (c) makes the important point that larger sized windows have inherently a better chance of predicting well the curvature at the center of the window. In (c) the green window is of size 3x3, while in (b) it is of size 5x5. The underlying discrete labelling is identical in both cases and the red curve is the optimal (lowest curvature) continuous curve given the window.
%The crucial point is that the curvature of the central pixel of the continuous curve is very different in (b) and (c).
The crucial point is that the curvature of the continuous curve, at the center
of the window, is very different in (b) and (c).
Note, this problem is to some extend mitigated by the fact that the total cost of segmentation is the sum of costs along the boundary.}
\label{fig:car_curve}
\end{center}
\vspace{-4mm}
\end{figure}

Suppose we are given a shape $S\subset \Real^2$ such that we can calculate the curvature $\kappa$ at every point of the boundary, $\partial S$. Let $f(\kappa)\geq 0$ be a curvature cost function, which defines a desired penalty on curvature, in this paper we consider $f(\kappa) = \min(\kappa^2,f^{\rm max})$.
Let the total cost of the shape be $\int_{\partial S} f(\kappa) dl$. Our goal is to approximate this integral by the sum
\begin{equation}
\label{bndsum}
\sum_{h \in \U} E_h(\x),
\end{equation}
where functions $E_h$ operate over a discretized representation of the shape, $\x$, and
are of the form~\eqref{HOterm} with weights $\w, \bc$. Here $\w$ and $\bc$ denote the concatenated vectors of all weights $\w_{y}$ and $c_{y}$, respectively.
The learning problem is to determine the pattern weights $\w,\bc$ such that the approximation is most accurate. Since the mapping of continuous to discrete curves is a many-to-one mapping, we further formalize our exact goal in figure \ref{fig:car_curve}. In the figure we also motivate the important aspect that larger windows are potentially superior.

We first restrict the sum in~\eqref{bndsum} to take into account only boundary locations. We call $h$ a {\em boundary location} for shape $\x$ if the $2{\times}2$ window at $h$ contains some pixels which are labeled foreground as well as some pixels which are labeled background, as illustrated in fig.~\ref{fig_bnd_shema}.
We constrain all soft patterns to be non-negative ($\<w_y,x\> + c_y \geq 0$) and introduce two special patterns (fig.~\ref{fig_bnd_shema}b,c), which have cost 0 for locations where the $2{\times}2$ window at location $h$ contains only background or foreground pixels. These patterns make $E_h(\x)$ vanish over all non-boundary locations, therefore such locations do not contribute to the sum~\eqref{bndsum}. The learning task is now to determine $E_h(\x)$, such that at each boundary location the true cost $f(\kappa)$ is approximated. In this way~\eqref{bndsum} does correspond to the desired integral if we were to neglect the fact that the number of boundary locations does only approximate the true length of the boundary. Note, the number of boundary locations does correspond to the ``Manhattan'' length of the boundary. We will come back to this problem in sec.~\ref{experiments}.
\paragraph{Point-wise learning procedure.} Let us assume that in a local $K{\times}K$ window, shapes of low curvature can be well-approximated by simple quadratic curves\footnote{Note, based on our definition in fig. \ref{fig:car_curve} we select quadratic curves which are likely to be the ones of lowest curvature (among all curves) for the corresponding discrete labelling.}. The idea is to take many examples of such shapes and fit $E_h(\x)$ to approximate their cost.
%The truncation of the approximation is achieved by the implicit special pattern $(\w_0=0, c_0 = f^{\rm max})$.
We consider many quadratic shapes $(S^i)_{i=1}^N$ in the window $K\times K$ and derive their corresponding discretization on the pixel grid $(\x^i)_{i=1}^N$. Each continuous shape has an associated curvature cost $f^i = f(\kappa^i)$ at the central boundary location. We formulate the learning problem as minimization of the average approximation error:
\begin{equation}\label{pointwise_approx}
\begin{array}{l}
\arg\min_{\w,\bc} \sum_{i} |E_h(\x^i) - f^i|,\\
\st \begin{cases}
\w_0=0, c_0 = f^{\rm max}\\
E_h(x)\geq 0 \tab \forall x
\end{cases}
\end{array}
\end{equation}
where the first constraint represents the special implicit pattern $(w_0,c_0=f^{\rm max})$, which ensures that $E_h(\x)\leq f^{\rm max}$. The second constraint makes sure that cost is non-negative. It is important for the following reason: the formulation of the approximation problem does not explicitly take into account ``negative samples'', \ie labellings which do not originate from smooth curves, and which must have high cost in the model. However, requiring that all possible negative samples in a $K{\times}K$ window have high cost would make the problem too constrained. The introduced non-negativity constraint is tractable and not too restrictive. This problem appears difficult, since $E_h(\x^i)$ is itself a concave function in the parameters $\w,\bc$. We approach~\eqref{pointwise_approx} by a k-means like procedure with a specially constructed initialization: \\[0pt] %We call this learning {\bf algorithm A}:
\begin{minipage}[c]{\linewidth}
\parbox{\linewidth}{\center \bf Alg.~1. Iterative Factor Discovery}
\noindent{\bf Input}: $\x^i$, $f^i$, $\w,\bc$\\
\noindent{\bf Repeat till convergence or maximum iterations:}\\
{\bf 1.} For all training images $i$ find best matching patterns $y^i = \arg\min\limits_{y} [\<\w_y,\x^i\> + c_y]$

{\bf 2.} For all $y\in 1\dots N_P$ refit $(\w_y,c_y)$:\\
\begin{equation}\label{refit_step}
\begin{array}{ll}
%(\w_j, c_j) = & \min\limits_{\w_j,c_j} \sum\limits_{i| y^i = j} |\w_j^T x^i +c_j - f^i|\\
%& \vphantom{H^{H^H}_{H_H}} \st \sum_v \min\limits_{\w_j,c_j} w_{j,v} x_v+c_j \geq \min\limits_{i| y^i = %j} f^i
(\w_y, c_y) = & \arg\min\limits_{\substack{\w_y, c_y\\ \xi \vphantom{H_{H_H}}}\ } \sum\limits_{i| y^i = y} |\<\w_y,\x^i\> +c_y - f^i|\\
& \vphantom{H^{H^H}_{H_H}} \ \ \st
\begin{cases}
\xi_v \leq w_{y,v}\\
\xi_v \leq 0\\
\sum_v\xi_v +c_y \geq 0
\end{cases}
%\min\limits_{\x \in \{0,1\}^{K^2}} \<\w_y, \x\> +c_y \geq \min\limits_{i| y^i = y} f^i \ .
\end{array}
\end{equation}
\end{minipage}

The refitting step~\eqref{refit_step} is a linear optimization which can be solved exactly. The constraint in~\eqref{refit_step} is an equivalent representation of the constraint $\<\w_y,\x\>+c_y \geq 0$\ $\forall x$, imposed by~\eqref{pointwise_approx}.

%Note that it involves the constraint enforcing that no labeling of the pattern receive a cost which is below the target function $f$. %While objective~\eqref{pointwise_approx} ensures that good examples are well approximated,
%Indeed:
%\begin{equation}
%\min_x \sum_v w_{j,v}x_v = \sum_v \min(w_{j,v},0).
% \end{equation}
%This is important for the following reason. Our approximation $E_h$ is the lower envelope of all the pattern costs. If there existed an arbitrary labeling which has a low cost in one pattern then it would also have a low cost in $E_h$. Hence, this constraint prevents arbitrary labellings from having a low cost.
The initialization and results of applying this learning procedure are discussed in the sec. \ref{experiments}. 
\sectionskip
\section{Inference}\label{optimization}
\sectionskip

\begin{figure}[ht]
\begin{center}
\setlength{\figwidth}{0.4\linewidth}
\setlength{\tabcolsep}{10pt}
\setlength{\doublerulesep}{2pt}
\setlength{\fboxsep}{0.3pt}
\begin{tabular}{cc}
\includegraphics[width=0.45\linewidth]{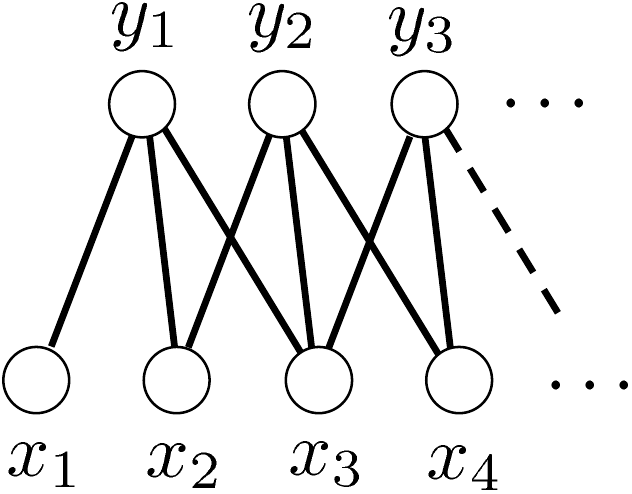}&
\includegraphics[width=0.38\linewidth]{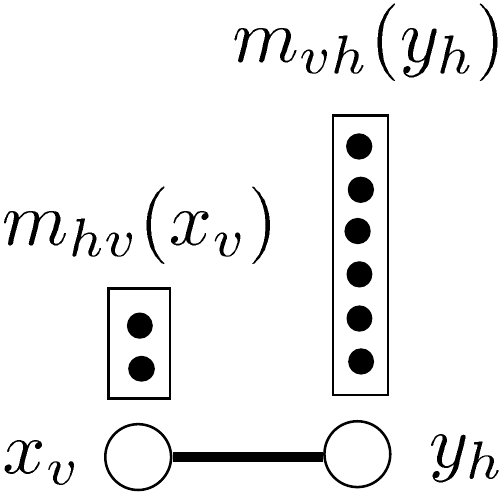}\\
(a) & (b)
\end{tabular}
\end{center}
\caption{
(a) Graphical model for the energy~\eqref{energy-pairwise}.
(b) Pairwise terms connecting pixel labeling $x_v$ and pattern-switching variable $y_h$. Circles show possible states: two states for $x_v$ and $N_P$ states for $y_h$. }
\label{fig_trws}
\end{figure}

We examined several standard MAP inference techniques for pairwise MRFs, and found the following two-stage procedure to work best for our problem.
First, the TRW-S algorithm\cite{Kolmogorov-06-convergent-pami,Wainwright03nips} is run for a fixed number of iterations. Then an initial solution is obtained by rounding the tree min-marginals. Second, a Block-ICM procedure improves on this initial solution.
%It is known that minimizing such energies is \cite{Kolmogorov-06-dense}.
\mypar{TRW-S}\label{trws}
In contrast to many other pairwise MRFs encountered in computer vision, our model has a very large number of pairwise links. We developed the following memory-efficient implementation of TRW-S for models with pairwise interactions between variables with few states and variables with many states.
%In contrast to many other pairwise MRFs encountered in computer vision, our model has a very large number of pairwise links. Hence, we had to modify the original TRW-S algorithm~\cite{Kolmogorov-06-convergent-pami} to make it more memory efficient.
%To be precise,
For an image of size $N_Y{\times}N_X$ and model with $N_{P}$ patterns of size $K{\times}K$, there are in total $O(N_Y N_X K^2)$ pairwise terms needed to represent the pattern potentials (pairwise energy~\eqref{energy-pairwise} can be illustrated as a bipartite graph in fig.~\ref{fig_trws}(a)). The algorithm in~\cite{Kolmogorov-06-convergent-pami} needs to keep a message (vector of size $N_{P}$) for each edge. This makes the original procedure~\cite{Kolmogorov-06-convergent-pami} extremely memory intensive. Since each pixel has only two labels and there are $N_P$ labels for the pattern switching variable, we can improve on memory requirement with a little computation overhead. We make the following modification to the implementation~\cite[fig.3]{Kolmogorov-06-convergent-pami}. %we can pick the order of variables such that
We store reparametrized unaries, $\hat\theta_s(x_s)$ and $\hat\theta_h(y_h)$, ($O(N_Y N_X N_P)$ memory) and messages only in the direction patterns$\rightarrow$pixels, $m_{hv}(x_v)$, ($O(N_Y N_X K^2)$ memory).
%Our implementation keeps only re-parameterized unaries, $O(N_Y N_X N_P)$ numbers, and messages of the form patterns$\rightarrow$pixels, $O(N_Y N_X K^2)$ numbers. When a reverse message is needed, it is directly computed from these values but not stored.
When the reverse message is requested by the algorithm it is computed on the fly using the equation
\begin{equation}
m_{vh}(y_h) = \min_{x_v}\big[ \gamma_{st} \hat\theta_s(x_s) - m_{hv}(x_v)+\theta_{vh}(x_v,y_h)\big],
\end{equation}
which is $O(N_P)$ computations. To completely specify the algorithm we have to choose the ordering of variables and the parameters $\gamma$. We can specify the ordering corresponding to longer chains, which potentially provide a faster convergence or the ordering where all $x$ precede all $y$, corresponding to short 1-edge chains, which makes the computation paralelizable. The parameters $\gamma$ are selected following the recommendation in~\cite{Kolmogorov-06-convergent-pami}.
\par
This modified version of TRW-S requires only $O(N_Y N_X$ $(K^2+N_P))$ memory and runs in $O(N_Y N_X K^2 N_P)$ time per full-pass iteration. In practice this results in about 5 seconds per iteration for an image of size $158{\times}128$, however a significant number of iterations was required to archive good results (e.g. up to 4000), which is a known issue for dense graphs~\cite{Kolmogorov-06-dense}.
\mypar{Block-ICM}
It often happens that the tree min-marginals of TRW-S for some pixels are ``indecisive'', i.e. possibly many different labellings have a low energy. In this case the solution picked by our pixel-independent rounding schemes may be rather poor. We found that further local improvements can significantly decrease the energy. Block-ICM tries to improve the current labeling $\x$ by switching states of a small block of $k$ variables at a time. Obviously, its complexity grows exponentially in $k$, hence $k$ must be low (we use k=6). During Block-ICM, the blocks are selected densely around the current boundary of $\x$. For an image of size $158{\times}128$ Block-ICM needs about 3 minutes to converge.
\mypar{LP relaxation}
\begin{figure}[tr]
\begin{center}
\setlength{\tabcolsep}{2pt}
\setlength{\doublerulesep}{0pt}
\setlength{\fboxsep}{0pt}
%\input{}
%\fbox{
\begin{tabular}{ccc}
\fbox{\includegraphics[width=0.27\linewidth]{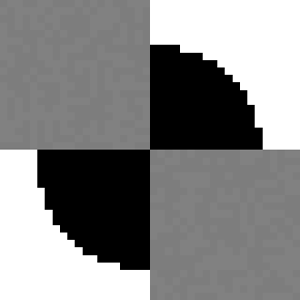}}&
\fbox{\includegraphics[width=0.27\linewidth]{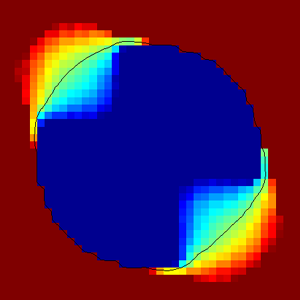}}&
\fbox{\includegraphics[width=0.27\linewidth]{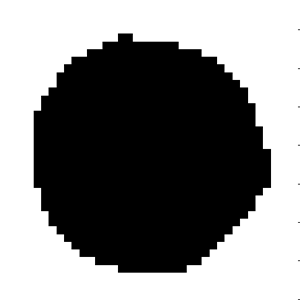}}\\
 & {\small LB = 1.169} & {\small E = 3.4} \\
(a) & (b) & (c) \\[2mm]
%\vspace{2mm}\\
\fbox{\includegraphics[width=0.27\linewidth]{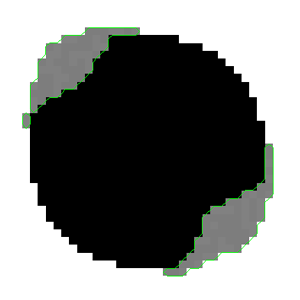}}&
\fbox{\includegraphics[width=0.27\linewidth]{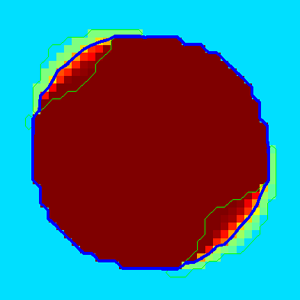}}&
\fbox{\includegraphics[width=0.27\linewidth]{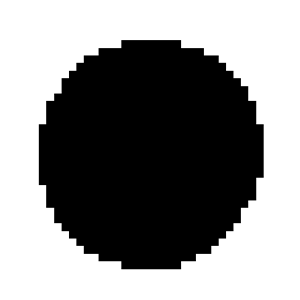}}\\
 & {\small LB=1.25} & {\small E = 2.06} \\
(d) & (e) & (f)\\[2mm]
%\vspace{2mm}\\
\end{tabular}
%\includegraphics[width=\linewidth]{fig/inpaint/circle_old.png_circle}%
%\includegraphics[width = \linewidth, height = \linewidth]{fig/dummy}
%}
\caption{ {\bf Restricted LP.} (a) Unaries: black -- foreground, white -- background, gray -- area to be inpainted;
(b) Tree min-marginals of TRW-S: a thin line shows the 0-level contour; (c) Rounding of TRW-S solution; (d) The reduced problem; (e) Relaxed primal solution of the reduced problem (with 0-level contour); (f) Rounding of relaxed primal solution. We also show lower bound (LB) for relaxation and energie (E) for a discrete solution. Note, the discrete, ground truth circle has an energy of $E=1.96$, which is still lower than the best solution found (f).}
\label{fig_inpaint_circle_lp}
\end{center}
\end{figure}

TRW-S is a suboptimal dual solver for linear program relaxation of the discrete pairwise energy minimization. The relaxation (see \eg~\cite{Werner-PAMI07} for an overview) is obtained by linearizing objective~\eqref{energy-pairwise} and dropping the integrality constraints. In particular, it replaces binary variables $x_v\in \{0,1\}$ with relaxed variables $\bar x_v\in [0, 1]$. The optimal relaxed labeling may be of interest even when the relaxation is not tight. However, TRW-S does not compute the primal relaxed labeling and it may get stuck in a suboptimal point. On the other hand solving the full primal LP seems infeasible since it requires $O(N_Y N_X K^2 N_P)$ variables.
%In our technical report \cite{TR} we show that the following heuristic procedure may improve the rounded solution of TRW-S, however, due to computational complexity we decided not use it in any experiment.
The following heuristic can be applied. Note, although, as we will see, the results may be improved with this heuristic, we did not use it in our experimental section due to its computational complexity.
The procedure is that we use the dual solution of TRW-S to greedily fix a larger part of the primal relaxed variables (both corresponding to pixel labels $\x$ and patterns $\y$). When the tree min-marginal for a label in a pixel is above a threshold compared to the minimal tree min-marginal in the pixel, we fix the corresponding primal variable to 0 and eliminate it. This gives a restricted linear program, which then can be solved by a primal method. Example in fig.~\ref{fig_inpaint_circle_lp} illustrates these steps for a circle inpainting problem of the type fig.~\ref{fig_inpaint_circles}. We make two observations: first, the restricted LP attains optimality in a ``fractional'' relaxed solution, so there exist an integrality gap and we can not obtain an optimal discrete solution. Second, the value of the objective at the optimum of restricted LP is different from the lower bound by TRW-S, which means that TRW-S has converged to a sub-optimal dual solution.
%To each primal relaxed variable $\bar x_v$ there correspond a dual constraint $z_v \leq \psi_v$
%We solve the restricted LP, where for tree-min-marginals which are above a threshold from being minimal we restrict primal relaxed variables to be

Let us also briefly mention on the performance of Belief Propagation (BP), which we found inferior. We used the variant called sequential (min-sum) Belief Propagation (BP), obtained from TRW-S by setting all $\gamma$ to 1 as described in~\cite{Kolmogorov-06-convergent-pami}. In~\cite{Rother09} it was reported that BP performs best for texture denoising with soft pattern-potentials, while TRW-S performed poorly. For our model we observed the opposite: BP may produce very poor results, especially for problems where there is a large area of uncertainty (shape-inpainting). Note, we also tried dumping and different reordering heuristics for BP, but without success. We believe that an interesting direction for future work is to thoroughly compare various optimization schema for various types of soft pattern-based potentials.
\sectionskip
\section{Experiments}\label{experiments}
\sectionskip
\begin{figure*}[tr]
\centering
\setlength{\figwidth}{0.33\linewidth}
\setlength{\tabcolsep}{0pt}
\setlength{\doublerulesep}{0pt}
%\input{}
%\fbox{
\begin{tabular}[t]{ccc}
\includegraphics[width=\figwidth]{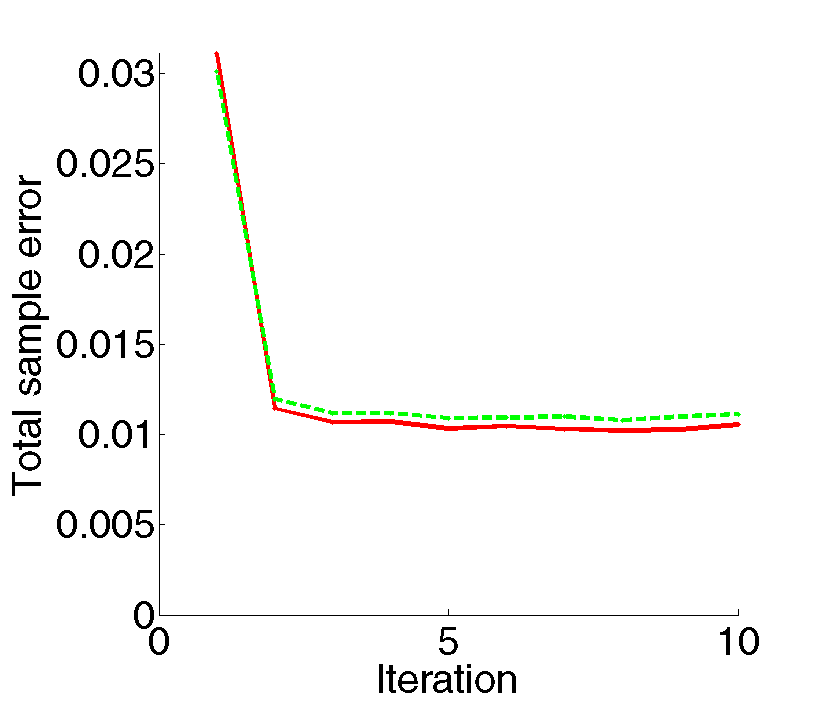}&
\includegraphics[width=\figwidth]{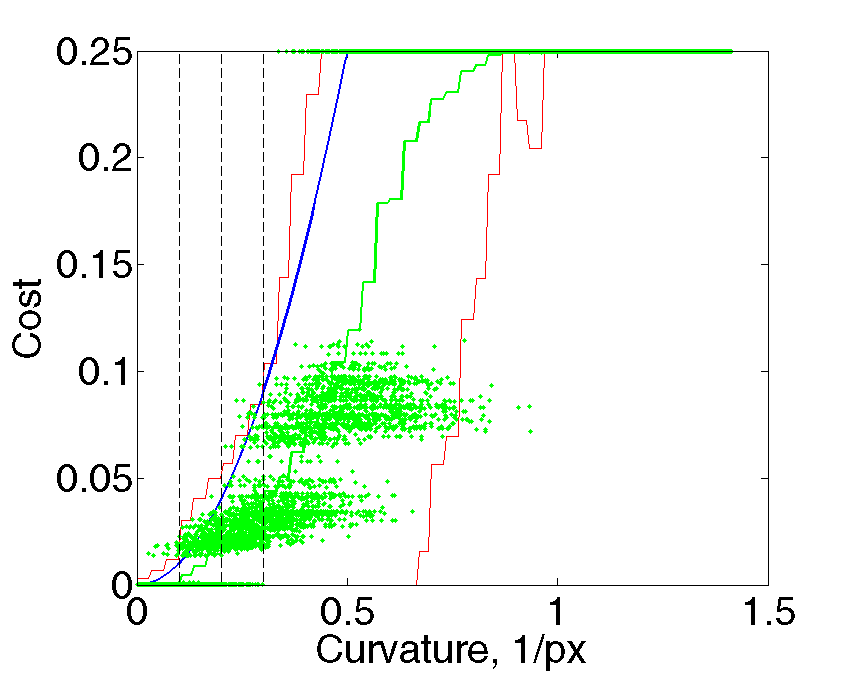} &
\includegraphics[width=\figwidth]{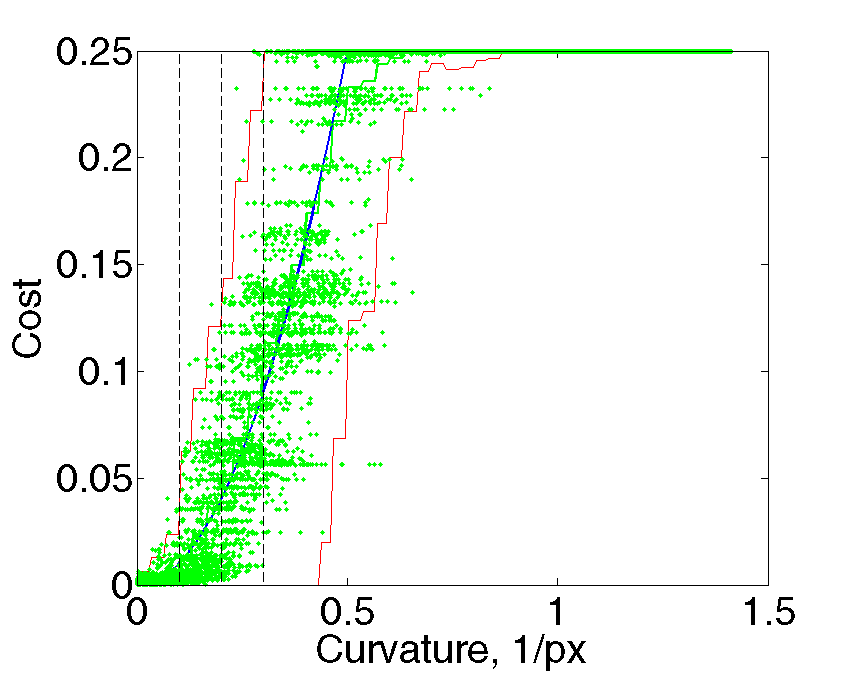}\\
(a) & (b) & (c)\\
\end{tabular}
\begin{tabular}{ccc}
\begin{tabular}{c}
\includegraphics[width=\figwidth]{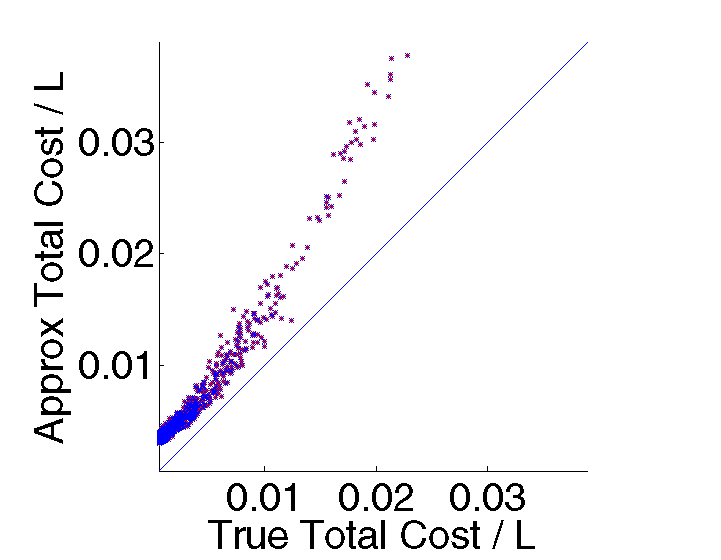}\\
\end{tabular}
\hspace{-10mm}
(d)\ \
&
\begin{tabular}{c}
\includegraphics[width=\figwidth]{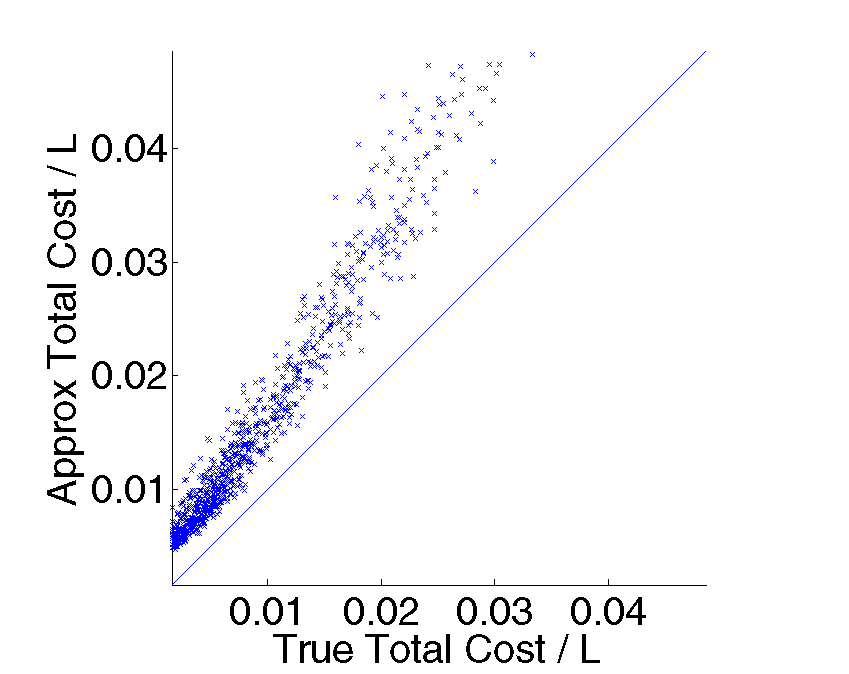}\\
\end{tabular}
\hspace{-10mm}
(e)\ \ \ \ \ \ \ \ \ 
&
\begin{tabular}{c}
\includegraphics[width=0.75\figwidth]{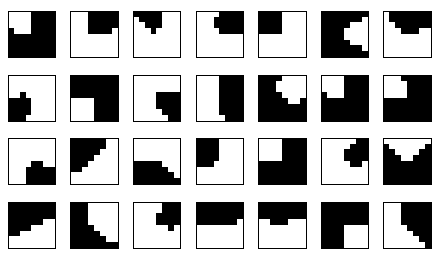}\\
(f)\\
%\includegraphics[width=1\figwidth]{fig/sample_circles_inline}\\
%(g)
\end{tabular}\\
%(d) & (e) & \\
\multicolumn{3}{c}{
(g) \begin{tabular}{c}\includegraphics[width=2.8\figwidth]{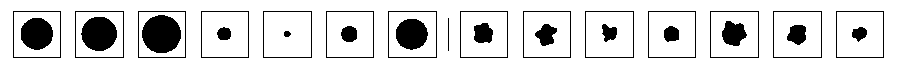}\end{tabular}%
}
%} & \multirow{2}{*}{ (g) \hspace{0.8\figwidth}}\\
%\multicolumn{2}{c}{
%\includegraphics[width=2\figwidth]{fig/sample_fourier_inline}
%}\\
%
\end{tabular}
%\includegraphics[width=2\figwidth]{fig/sample_circles_inline}\\
%\includegraphics[width=2\figwidth]{fig/sample_fourier_inline}\\

%\includegraphics[width=0.5\linewidth]{fig/dummy}%
%\includegraphics[width=0.5\linewidth]{fig/dummy}%
%\includegraphics[width = \linewidth, height = \linewidth]{fig/dummy}
%}
\caption{{\bf Cost Approximation.}
(a) The performance of Alg. 1 measured in terms of the objective~\eqref{pointwise_approx}. The red and green dashed curves show training and test error respectively. (b) Point-wise approximation cost with initial patterns, and (c) point-wise approximation cost after 10 iterations of Alg 1.
In both (b,c): each green point is a test sample; the blue line shows the desired true cost; the green line the mean approximation cost; and the red lines show $3{\times}$standard deviation bounds.
(d,e) Approximate total cost / length vs. true total cost / length for circles (d) and Fourier shapes (e).
(f) Examples of training and test patches used in (a-c).
(g) Examples of discretized shapes for circular shapes and Fourier shapes used in (d,e).
}
\label{fig_pointwise}
\end{figure*}
We performed several kinds of experiments.
\par\noindent 1. We discuss the learning procedure of our model and investigate the approximation quality it achieves. For the later we generated continuous shapes, for which the true total cost can be computed precisely, and compared that to our model. Note, to make inference and learning feasible we can only use a restricted number of patterns with limited size. We show that this gives a reasonable approximation of the desired cost functional. Note, in theory, by increasing the resolution (size of patterns) and the number of patterns one may archive approximation with an arbitrary accuracy.
\par\noindent 2.
The second set of experiments studies the task of ``shape-inpainting'' where the optimal segmentation (shape) has to be inferred, while only a few boundary conditions are given. It provides a good way of inspecting our prior shape model and assessing whether it corresponds to our intuitive notion of natural shapes.
\par\noindent 3.
The next experiment is on standard interactive image segmentation. Here we compare length versus curvature regularization.
\par\noindent 4.
Finally, we analyze the properties of the curvature model of~\cite{Schoenemann09} and provide some comparison with our model.

\mypar{Cost approximation} To learn our curvature model we used 96 patterns of size $K=8$. For the learning we sampled $N=10000$ random quadratic curves passing close to the center in a $K{\times}K$ pixel patch (fig.~\ref{fig_pointwise}(f)). The initial model is build as follows. We split the training patches into $32$ orientations, using the tangent of a curve, and also $3$ different curvature intervals. This gives in total 96 bins.
For each bin we fit a separate linear function using step 2 of Alg.1, which results in an initial set of 96 patterns.
We then run several iterations of Alg.1. The estimated error of the objective~\eqref{pointwise_approx} is shown in fig.~\ref{fig_pointwise}(a). We see that both  training and test error decrease over time.

\par
The initial and final point-wise cost (i.e. for a patch) is illustrated in Fig.~\ref{fig_pointwise} (b) and (c) respectively.
It can be seen that the mean of the approximated cost is very close to the true cost function, however, the variance is considerably large.
However, this problem should be mitigated when the cost is summed up along the full boundary, as the errors average out. To verify this, we sampled larger shapes for which we can compute the true total cost exactly and then compared to that of our model. Fig.~\ref{fig_pointwise}(d,e) shows this experiment for two classes of shapes: (d) circular shapes of size $100{\times}100$ with random radius (uniformly sampled in [5 50] pixels) and subpixel shift; (e) complex shapes created using Fourier series $\rho(\alpha) = a_0 + \sum_{k=1}^5 a_k \sin(k \alpha)+b_k \cos(k \alpha)$ in polar coordinates with random coefficients $(a,b)$. Derivatives, curvature and the total cost integral can be computed accurately for these shapes. We then measure the approximation error relative to the true length of the curve.
Figure~\ref{fig_pointwise}(d,e) shows that the variance is reduced, especially for shapes with low average curvature which are pre-dominant.
The plots do also reveal the fact that we consistently overestimate the true curvature cost. This problem is related to the fact that we approximate the integral along the boundary by the sum over boundary locations, which corresponds to the ``Manhattan'' length, which is usually higher than the Euclidean length. While this is not essential for getting a useful model for the shape prior, we discuss in Appendix~\ref{overlap} a way of reducing this error by adjusting the pattern costs.
%This experiment verifies that the approximation makes sense. We do not claim that the accuracy is comparable with state-of-the-art techniques of approximation curvature of digital curves, however such comparison could be informative in the future.
%
%\input{tex/fig_alg1_progress.tex}
%
%
%\input{tex/fig_total_cost.tex}
%
%
%
%
\mypar{Shape Inpainting}
The goal is to reconstruct the full shape, while only some parts of the shape are visible to the algorithm. This is a useful test to inspect our shape prior. Let $F\subset \V$ be the set of pixels restricted to foreground (shape) and $B\subset \V$ pixels restricted to background.
The unary terms of~\eqref{energy}, $\theta_{v}(x_v)$, are set to $\infty$ if label $x_v$ contradicts to the constrains and $0$ otherwise. This ensures that the correct segmentation is inferred in the region $F\cup B$.
%The unary terms of~\eqref{energy}, $\theta_{v}(x_v)$, are set accordingly to $\infty$, in order to ensure that the correct segmentation is inferred in the region $F\cup B$.
\begin{figure}%[tr]
\begin{center}
\setlength{\figwidth}{0.18\linewidth}
\setlength{\figwidtha}{0.10\linewidth}
\setlength{\tabcolsep}{2pt}
\setlength{\doublerulesep}{0pt}
\setlength{\fboxsep}{0.2pt}
%\input{}
%\fbox{
\begin{tabular}{lllll}
\fbox{\includegraphics[width=\figwidtha]{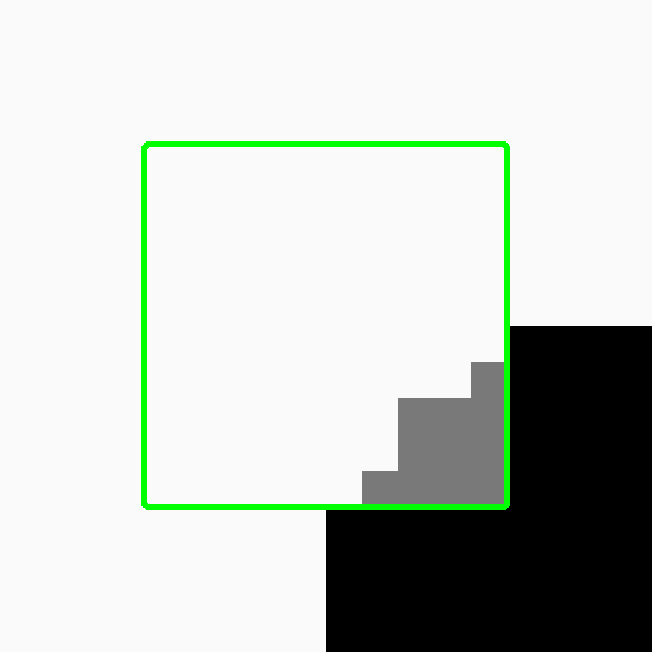}}&
\fbox{\includegraphics[width=\figwidtha]{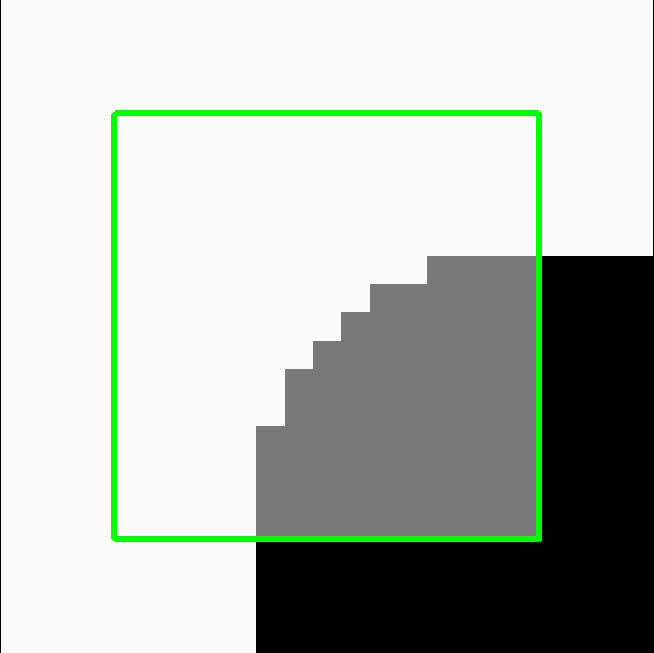}}&
\fbox{\includegraphics[width=\figwidtha]{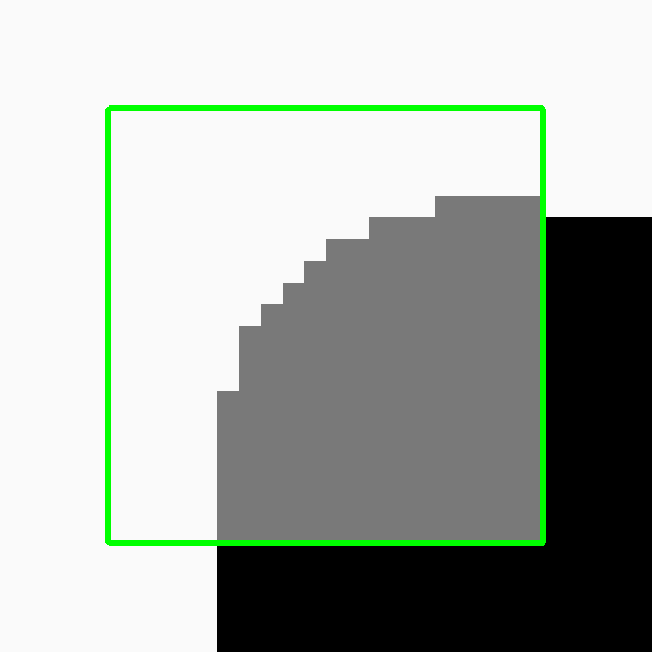}}&
\fbox{\includegraphics[width=\figwidtha]{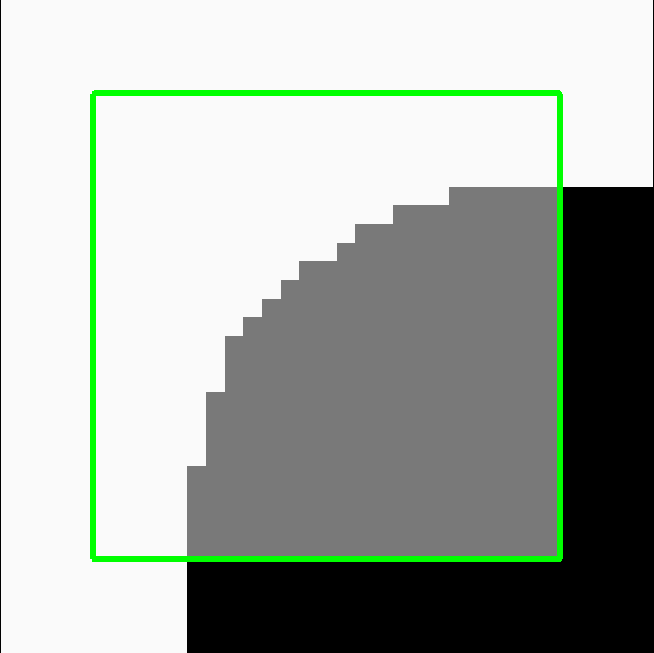}}&
\fbox{\includegraphics[width=\figwidtha]{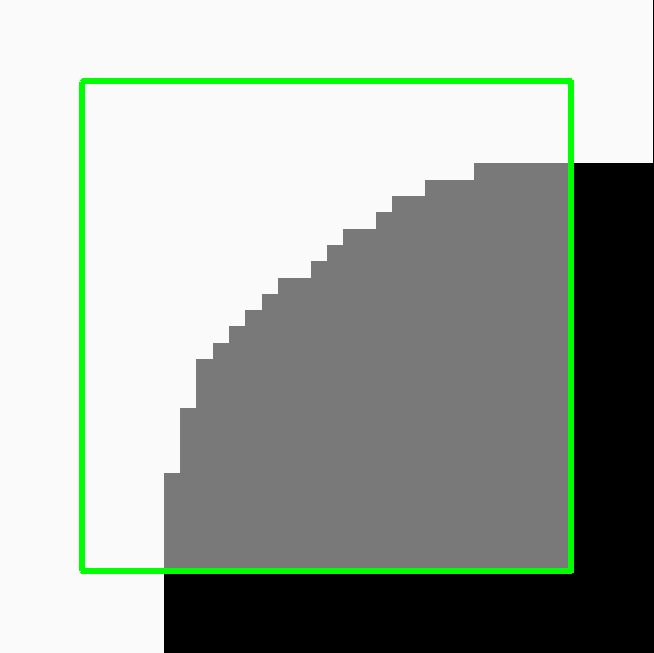}}\\
%(a) & (c) & (e) \\
\fbox{\includegraphics[width=\figwidth]{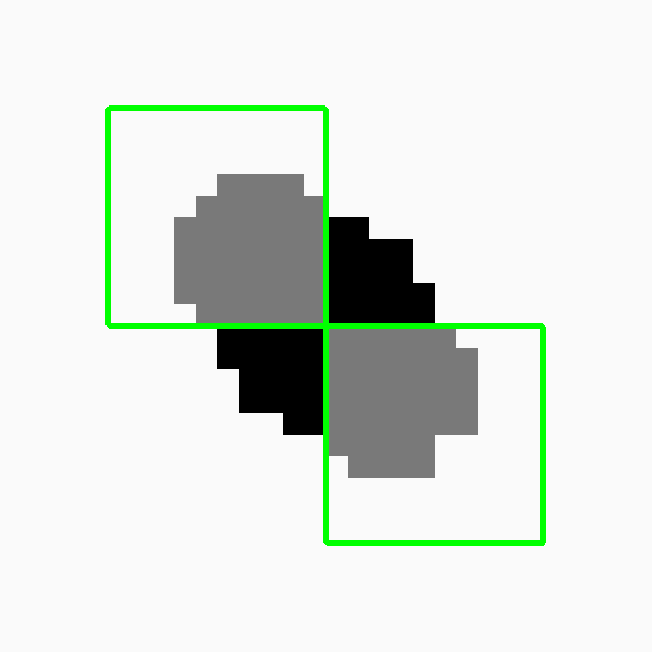}}&
\fbox{\includegraphics[width=\figwidth]{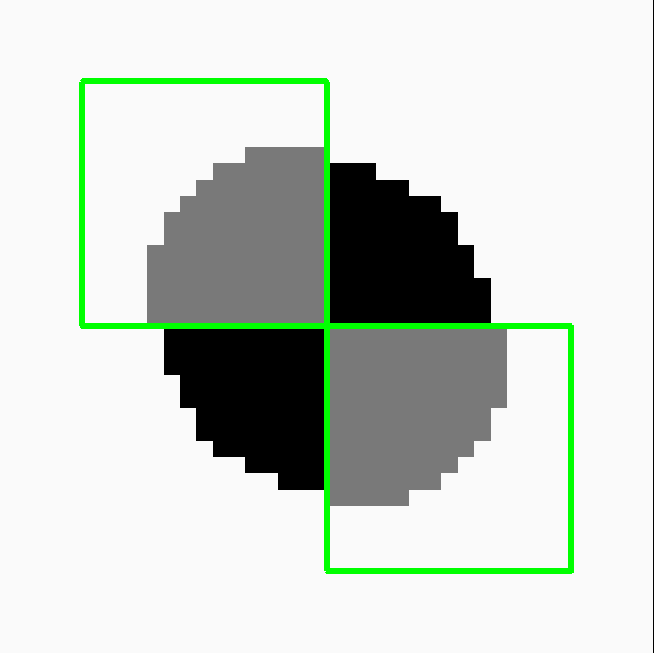}}&
\fbox{\includegraphics[width=\figwidth]{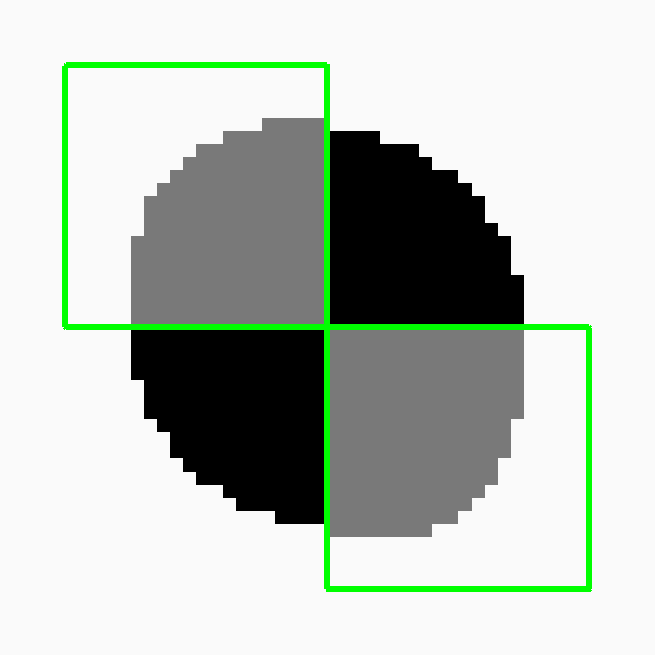}}&
\fbox{\includegraphics[width=\figwidth]{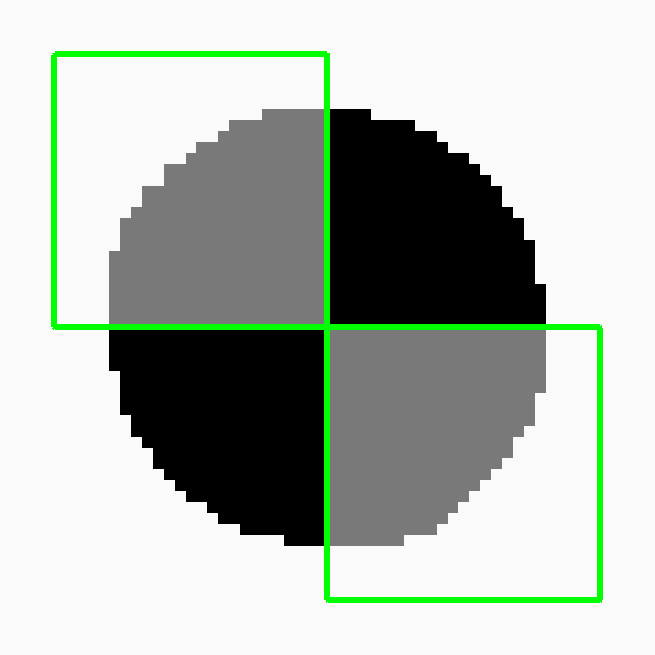}}&
\fbox{\includegraphics[width=\figwidth]{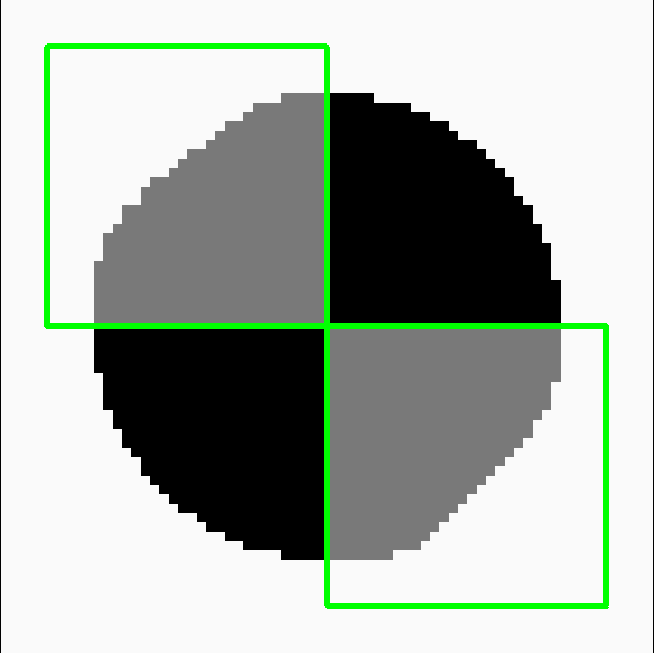}}\\
5 & 10 & 15 & 20 & 25\\
\end{tabular}
%
%\includegraphics[width = \linewidth, height = \linewidth]{fig/dummy}
%}
\end{center}
\caption{
Inpainting of a corner and a circle. The green boxes show the area to be inpainted, where the size in pixels of the length of green box is below the figures. Pixels in gray show the estimated solution. Note, the boundary conditions are different: right-angle boundary condition (top), circle boundary condition (bottom).
}
\label{fig_inpaint_circles}
\end{figure}
\begin{figure}%[tr]
\begin{center}
\setlength{\figwidth}{0.1\linewidth}
\setlength{\figwidtha}{0.1\linewidth}
\setlength{\tabcolsep}{1pt}
\setlength{\doublerulesep}{0pt}
\setlength{\fboxsep}{0.2pt}
%\input{}
%\fbox{
\tiny
\begin{tabular}{ccccccccc}
\tiny
\fbox{\includegraphics[width=\figwidtha]{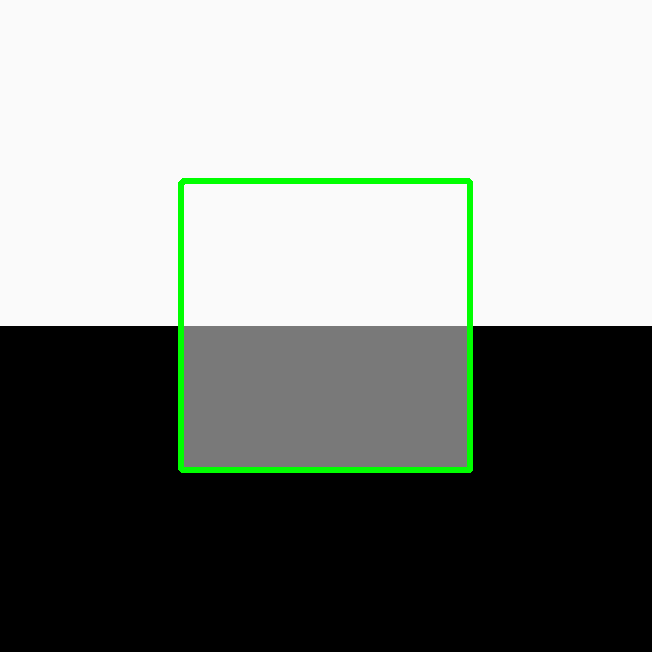}}&
\fbox{\includegraphics[width=\figwidtha]{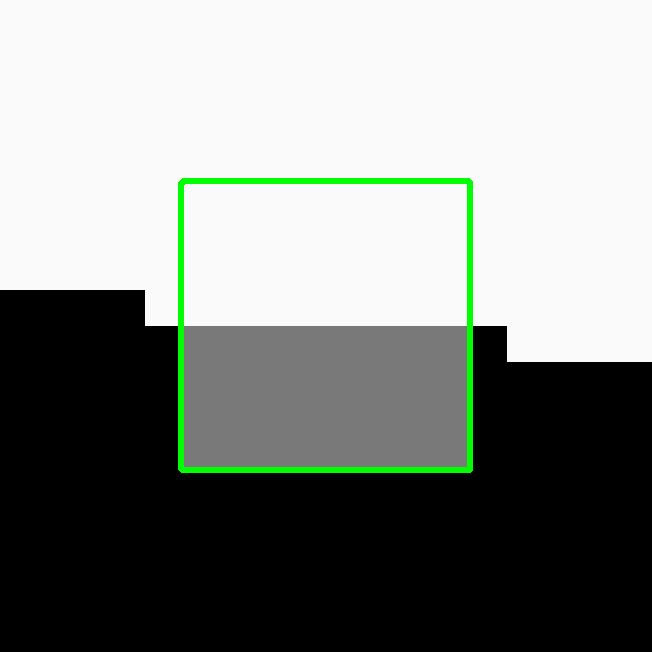}}&
\fbox{\includegraphics[width=\figwidtha]{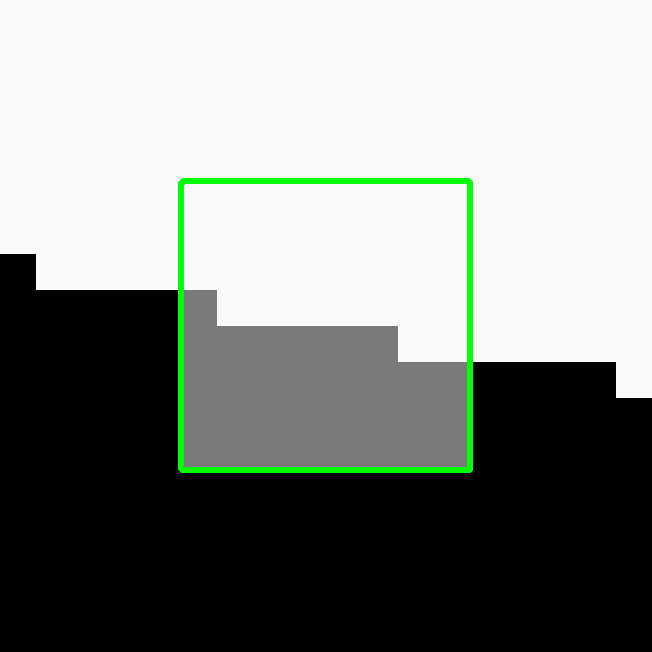}}&
\fbox{\includegraphics[width=\figwidtha]{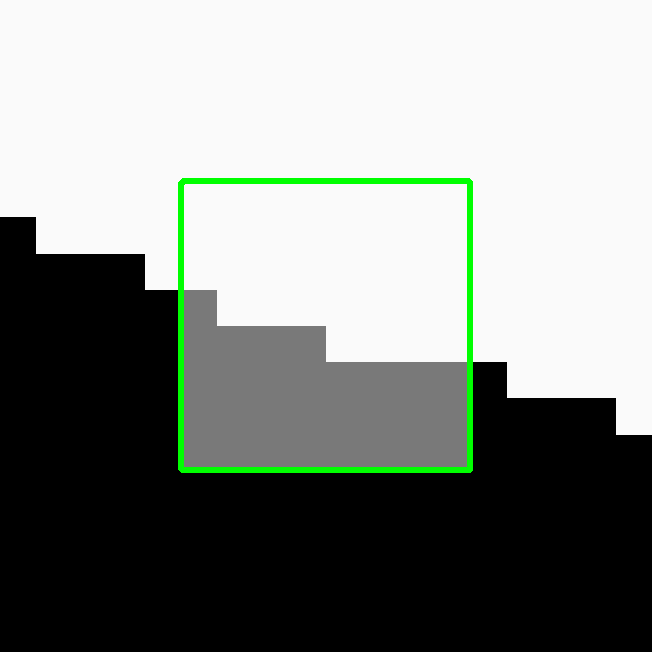}}&
\fbox{\includegraphics[width=\figwidtha]{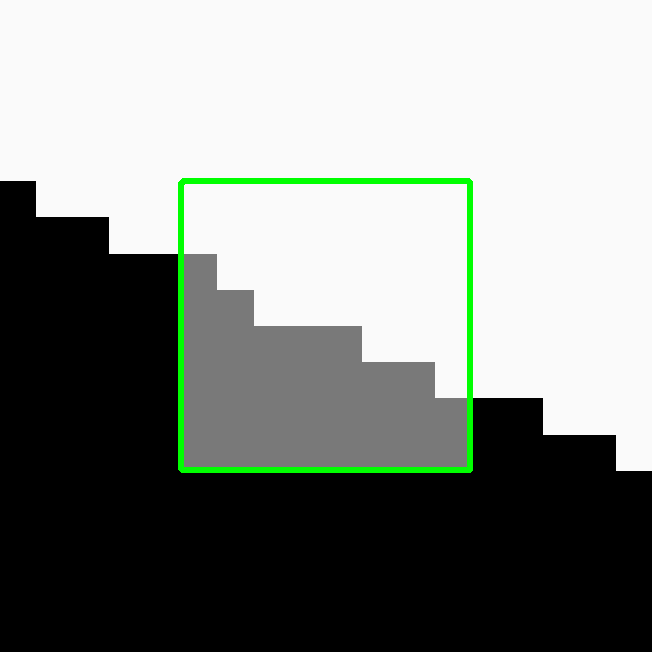}}&
\fbox{\includegraphics[width=\figwidtha]{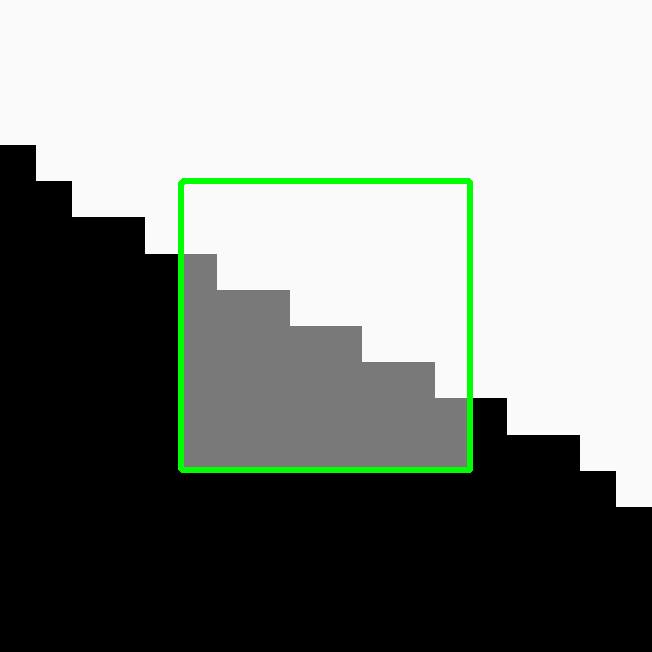}}&
\fbox{\includegraphics[width=\figwidtha]{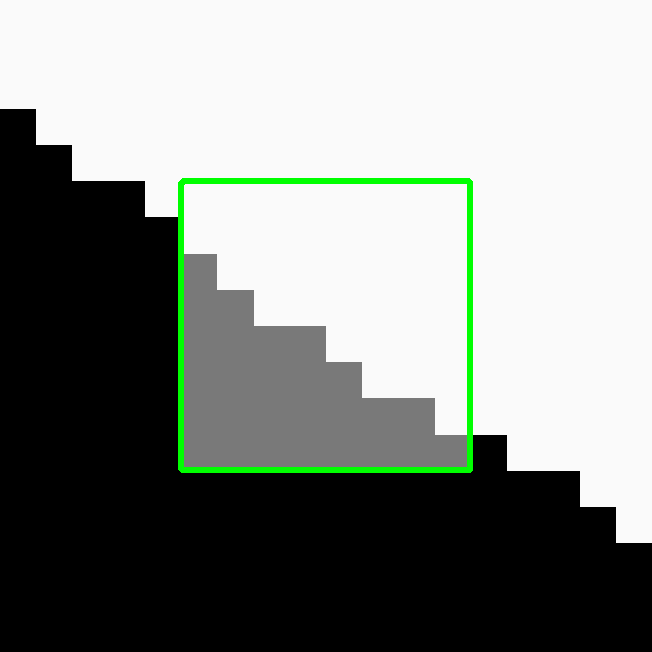}}&
\fbox{\includegraphics[width=\figwidtha]{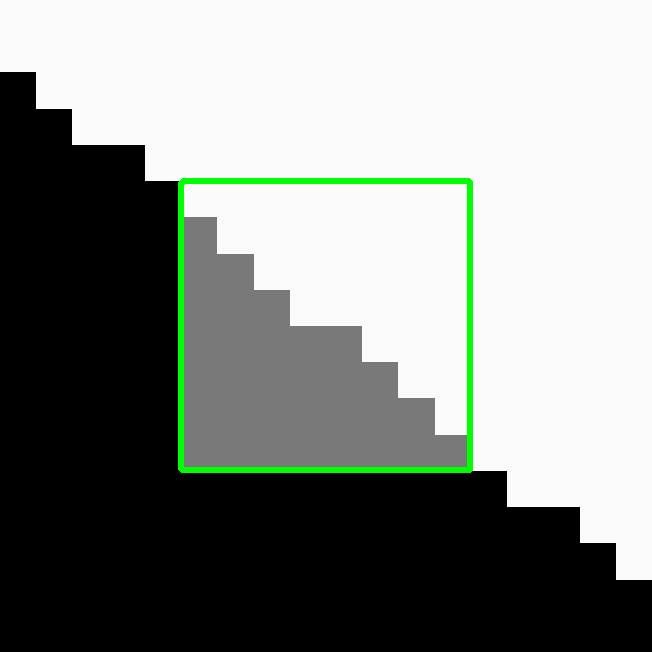}}&
\fbox{\includegraphics[width=\figwidtha]{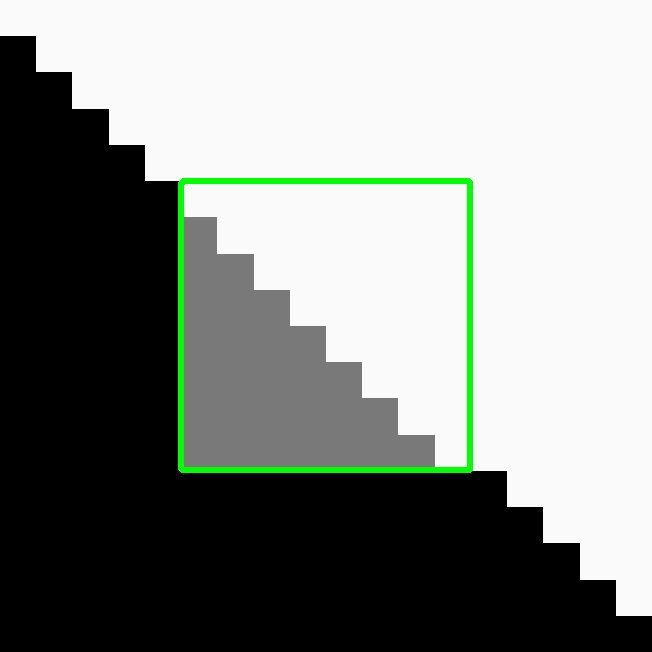}}\\
0.024  &  0.038  &  0.033  &  0.036  &  0.027  &  0.035  &  0.038  &  0.040  &  0.038\\
0.024  &  0.038  &  0.035  &  0.038  &  0.030  &  0.035  &  0.043  &  0.040  &  0.039\\
\fbox{\includegraphics[width=\figwidtha]{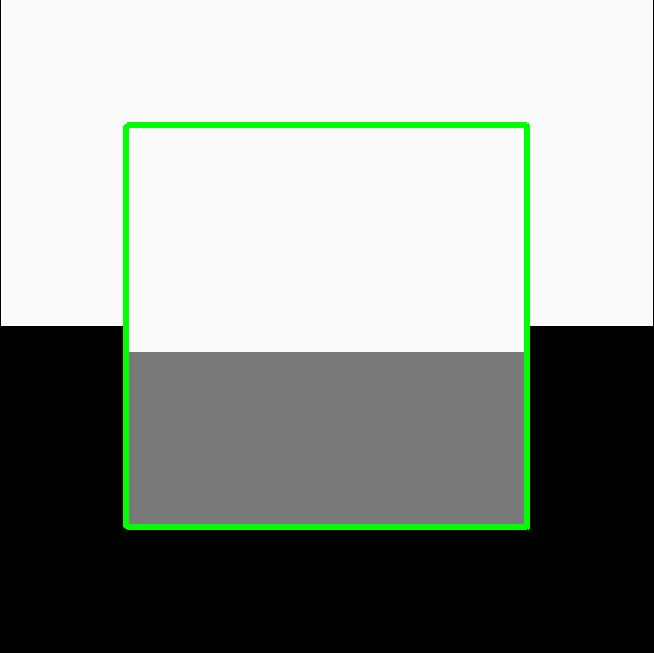}}&
\fbox{\includegraphics[width=\figwidtha]{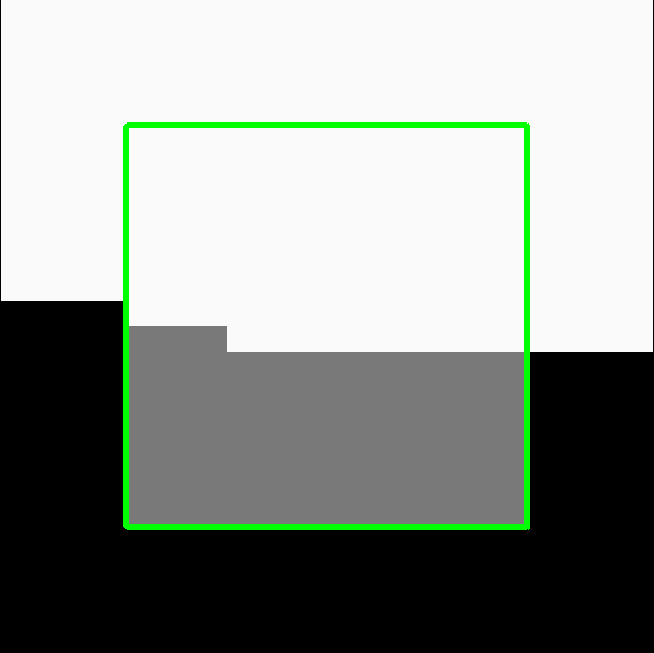}}&
\fbox{\includegraphics[width=\figwidtha]{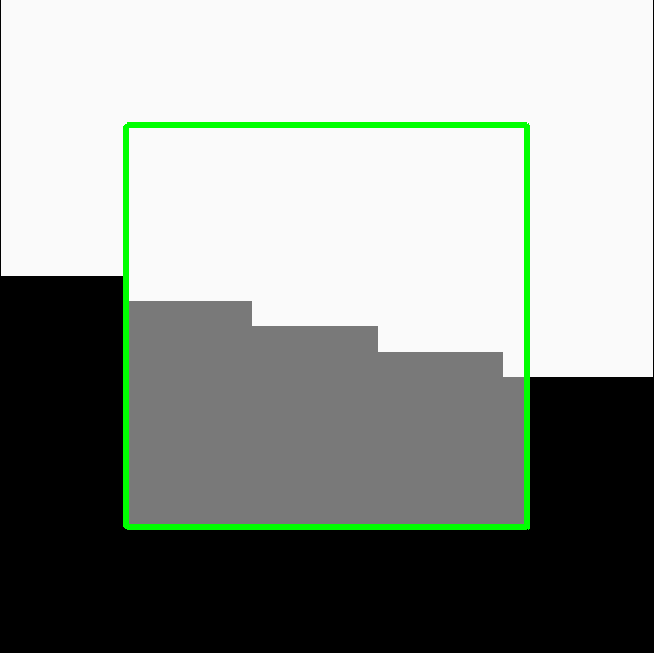}}&
\fbox{\includegraphics[width=\figwidtha]{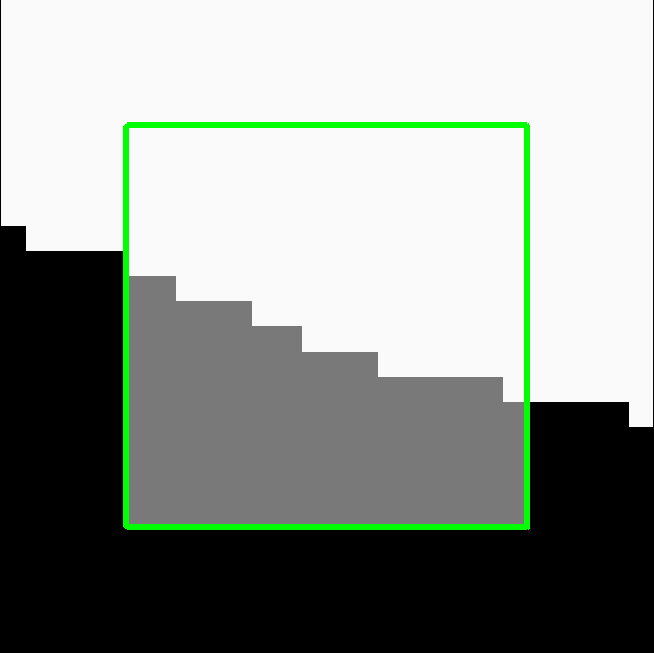}}&
\fbox{\includegraphics[width=\figwidtha]{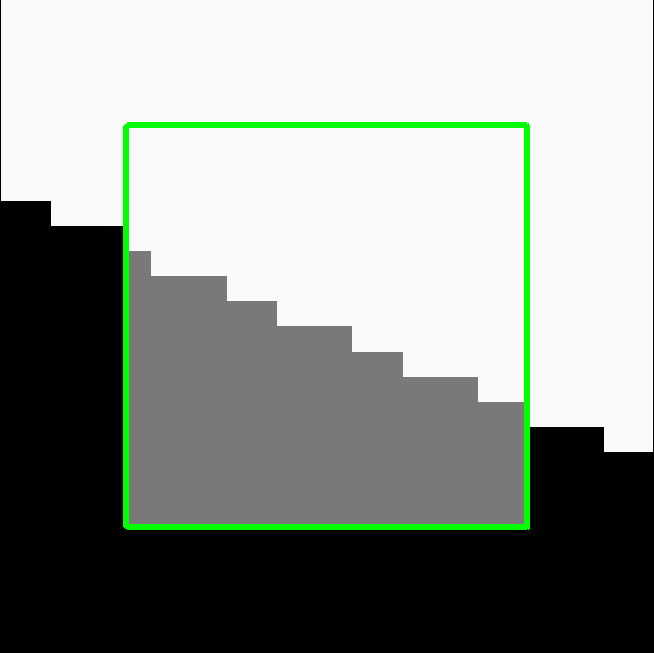}}&
\fbox{\includegraphics[width=\figwidtha]{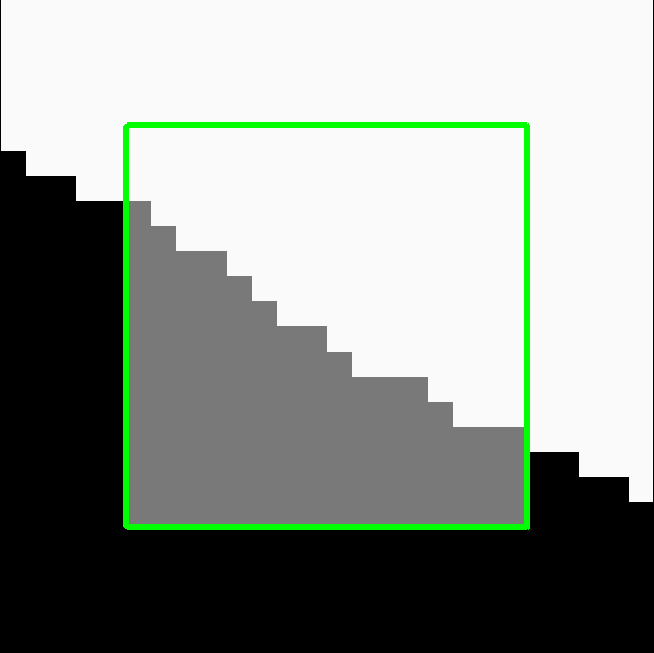}}&
\fbox{\includegraphics[width=\figwidtha]{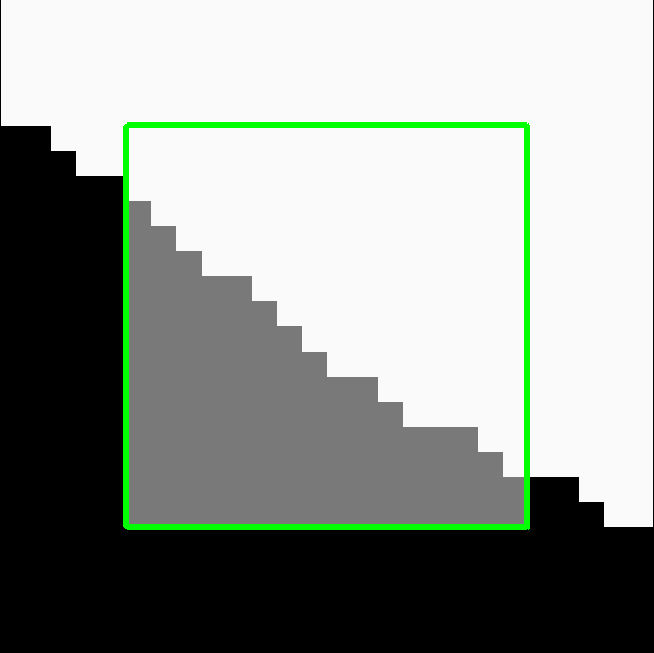}}&
\fbox{\includegraphics[width=\figwidtha]{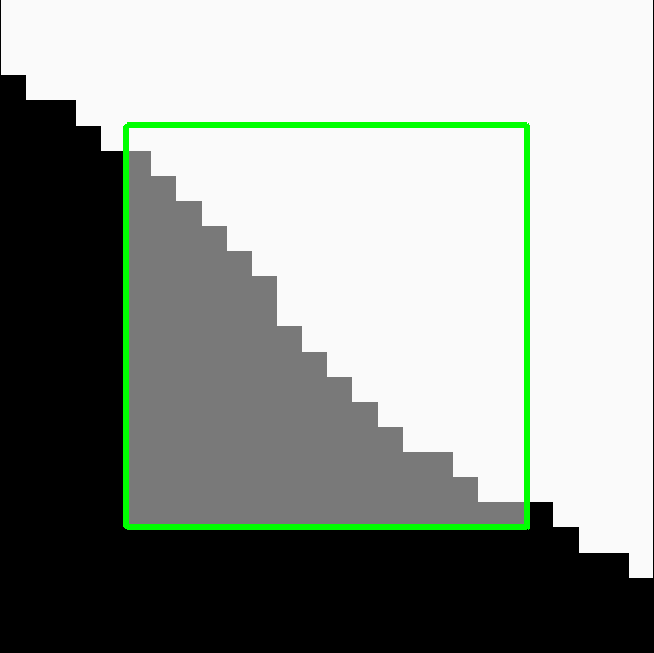}}&
\fbox{\includegraphics[width=\figwidtha]{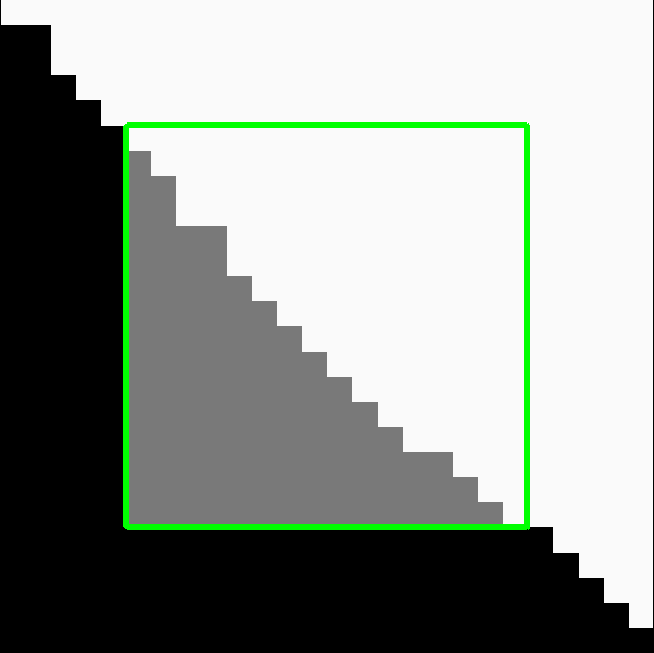}}\\
0.057  &  0.052  &  0.057  &  0.058  &  0.045  &  0.059  &  0.056  &  0.075  &  0.061\\
0.042  &  0.066  &  0.059  &  0.066  &  0.053  &  0.058  &  0.065  &  0.062  &  0.066\\
\fbox{\includegraphics[width=\figwidtha]{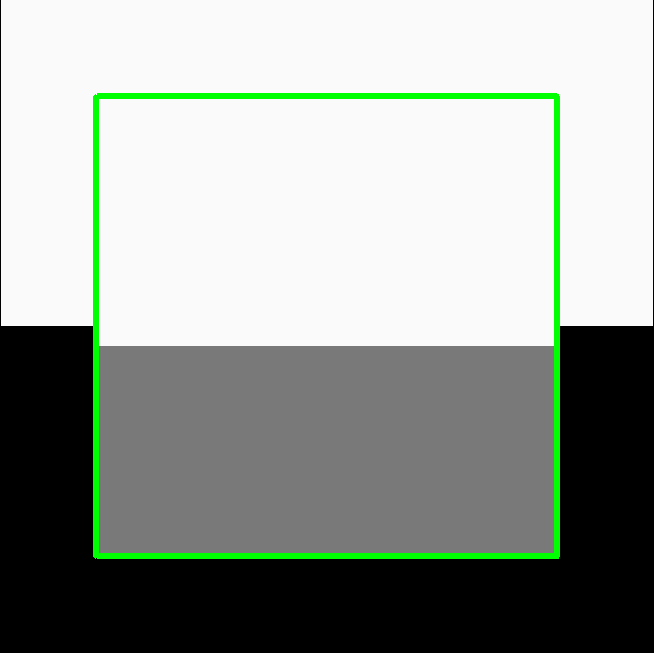}}&
\fbox{\includegraphics[width=\figwidtha]{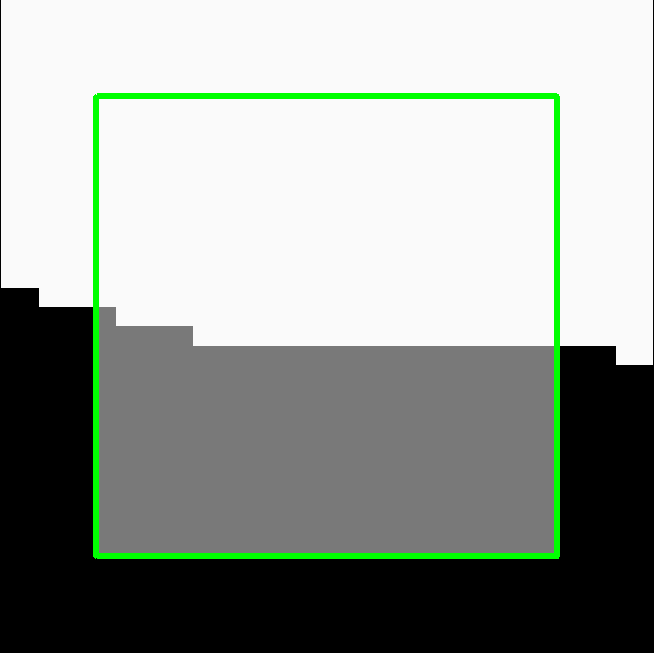}}&
\fbox{\includegraphics[width=\figwidtha]{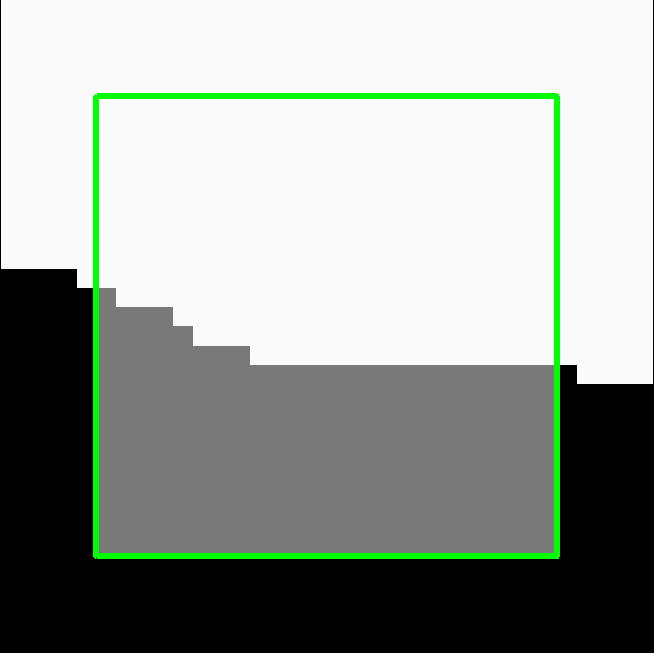}}&
\fbox{\includegraphics[width=\figwidtha]{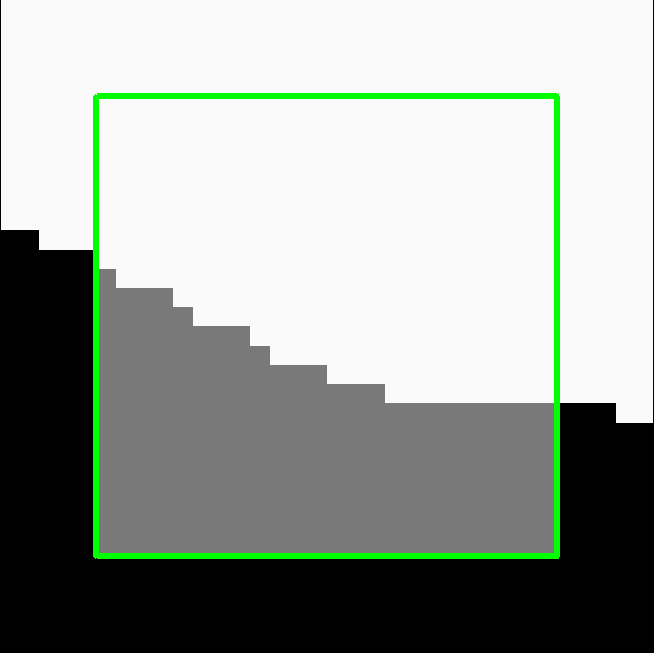}}&
\fbox{\includegraphics[width=\figwidtha]{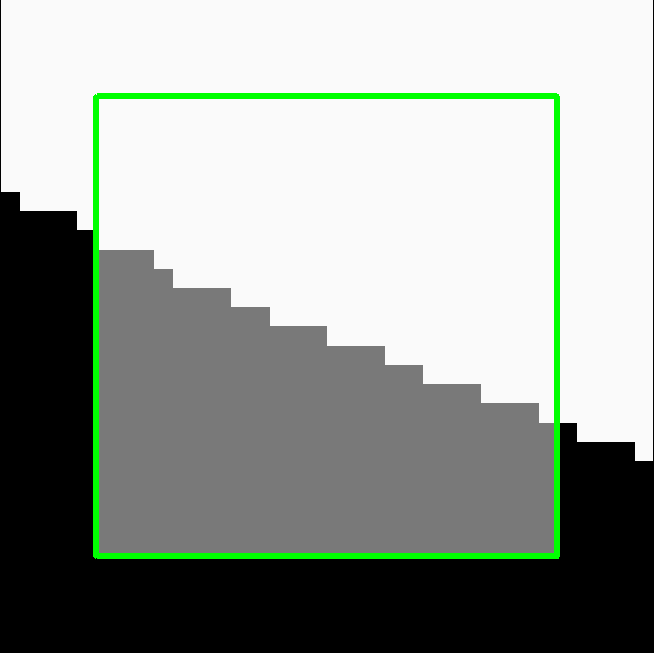}}&
\fbox{\includegraphics[width=\figwidtha]{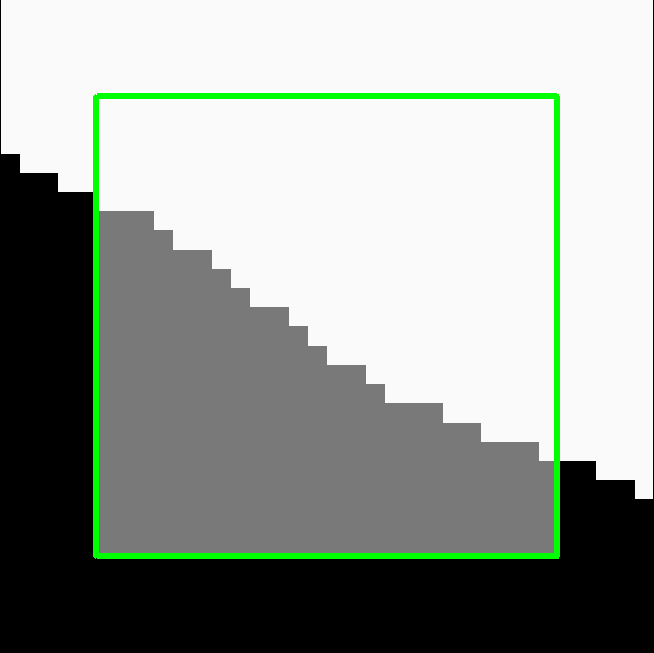}}&
\fbox{\includegraphics[width=\figwidtha]{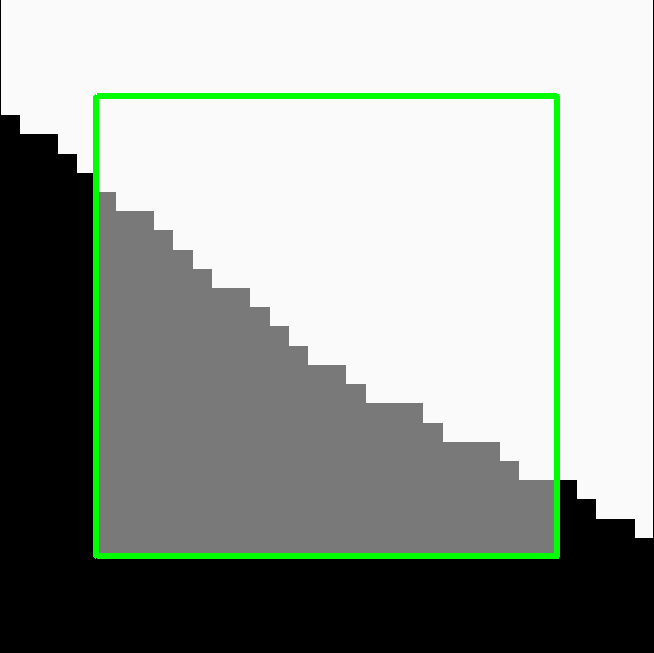}}&
\fbox{\includegraphics[width=\figwidtha]{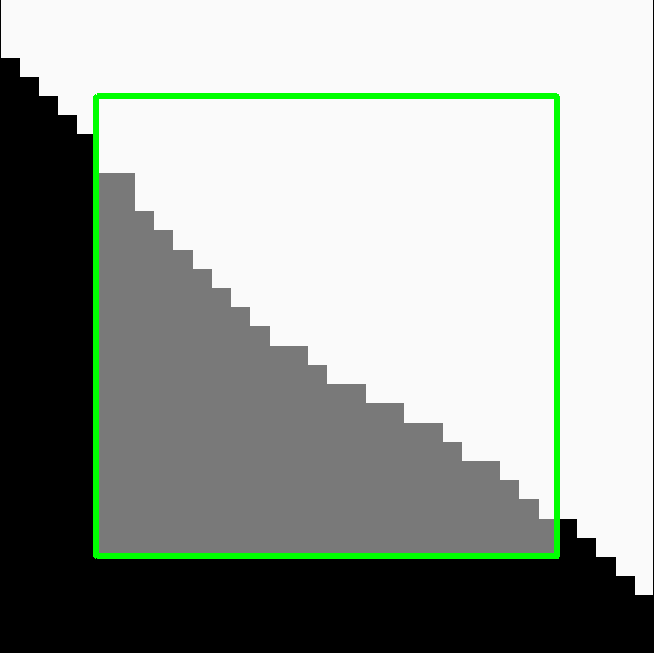}}&
\fbox{\includegraphics[width=\figwidtha]{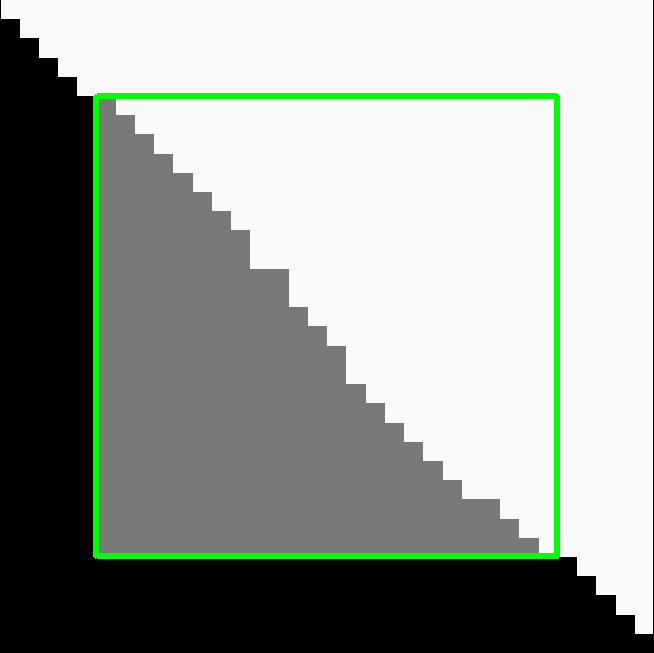}}\\
0.074  &  0.075  &  0.083  &  0.077  &  0.065  &  0.082  &  0.083  &  0.090  &  0.088\\
0.060  &  0.093  &  0.084  &  0.096  &  0.072  &  0.084  &  0.098  &  0.092  &  0.107\\
\end{tabular}
%
%\includegraphics[width = \linewidth, height = \linewidth]{fig/dummy}
%}
\end{center}
\caption{
Inpainting with a straight line boundary condition. The green box is of size 8, 16 and 24 pixels respectively from top to bottom. The numbers show the model cost of the estimated solution (top line) and the cost of the ground truth (discretized) straight line (bottom line).
%Inpainting of straight line boundaries. Top-to bottom a window of size 8, 16 and 24 px to be inpained. The numbers show the model cost of the displayed solution and that of the ground truth. It is seen that often the found solution is not approximating well the straight line while having lower energy. Clearly, this is drawback of the model and not of the optimization. The only case which might resulted as a drawback of optimization is shown in black.
}
\label{fig_inpaint_lines}
\end{figure}

%0.024    0.038    0.033    0.036    0.027    0.035    0.038    0.040    0.038
%0.024    0.038    0.035    0.038    0.030    0.035    0.043    0.040    0.039

%0.057    0.052    0.057    0.058    0.045    0.059    0.056    0.075    0.061
%0.042    0.066    0.059    0.066    0.053    0.058    0.065    0.062    0.066

%0.074    0.075    0.083    0.077    0.065    0.082    0.083    0.090    0.088
%0.060    0.093    0.084    0.096    0.072    0.084    0.098    0.092    0.107
In the unknown region $\V\backslash (F\cup B)$ all unaries are exactly $0$. Fig.\ref{fig_inpaint_circles},\ref{fig_inpaint_lines} show results for different inpainting problems with various boundary conditions corresponding to inpainting of some simple shapes. The main conclusion is that all results look reasonably good.
Note, for a tiny circle in fig.~\ref{fig_inpaint_circles}(bottom left) the reconstruction looks more like an oval than a circle. This is an expected result, and visually acceptable, since the boundary condition (black pixels) may correspond to either of the shapes and an oval has a lower cost. Also, we see that for some line inpainting examples in fig.\ref{fig_inpaint_lines} the result deviates slightly from the ground truth (straight line). Given that the cost of our solution is almost always lower than the cost of the ground truth, it is quite likely that the problem is due to a non-perfect model and not due to a local minimum of the optimization. As mentioned before, one way to overcome this problem is to allow for more patterns.
\begin{figure}[tr]
\begin{center}
\includegraphics[width=\columnwidth]{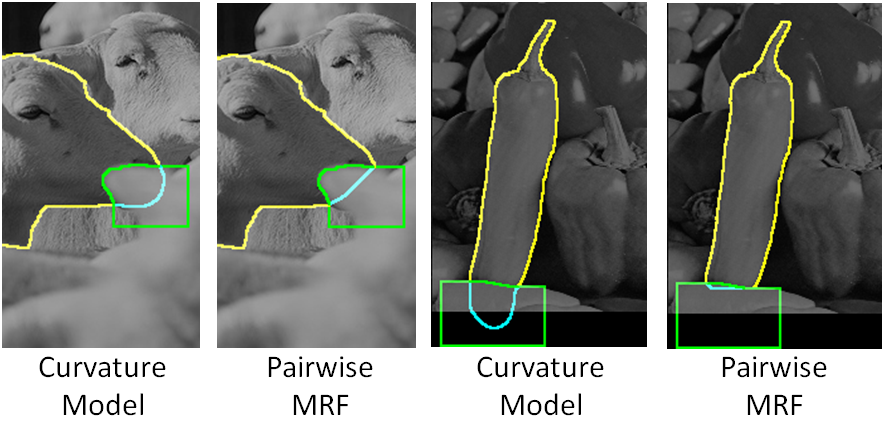}
\caption{\it Two example for automatic shape completions of an occluded object. In both cases the left result is with a pure curvature prior and the right result with a pure length prior (8-connected). Note, the yellow curve (and a part of the green curve) indicate the original user-defined segmentation. Then the user defines the green area. Inside the green area, the method automatically finds the shape completion (blue curve).}
\label{fig_inpaint_real}
\end{center}
\vspace{-4mm}
\end{figure}
We also tested our algorithm on inpainting real world images, and compared its results with those obtained by using a pairwise Markov Random Field formulation that tries to reduce the boundary length. The results can be seen in figures~\ref{fig_inpaint_real} and~\ref{fig:teaser}. It can be seen that the higher-order model that encodes curvature produces shape completions with smooth boundaries. An example combining curvature and length
priors is shown in fig.~\ref{fig_star_length}.
\begin{figure}%[tr]
\begin{center}
\setlength{\figwidth}{0.31\linewidth}
\setlength{\tabcolsep}{2pt}
\setlength{\doublerulesep}{0pt}
\setlength{\fboxsep}{0.2pt}
%\input{}
%\fbox{
\scriptsize
\begin{tabular}{ccc}
\fbox{\includegraphics[width=\figwidth]{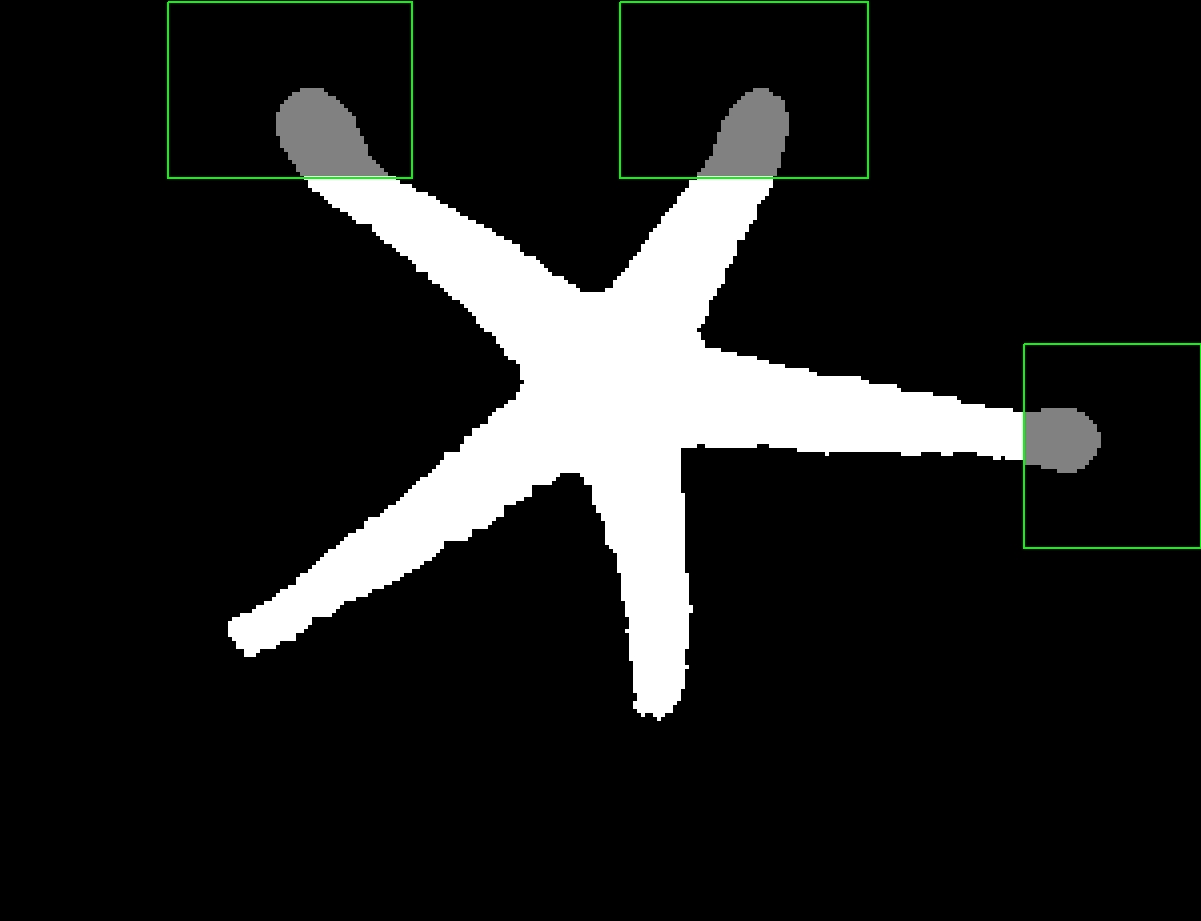}}&
%\fbox{\includegraphics[width=\figwidth]{fig/inpaint_real/star_inpaint2_small_display_K8_cvpr1a_cll}}&
\fbox{\includegraphics[width=\figwidth]{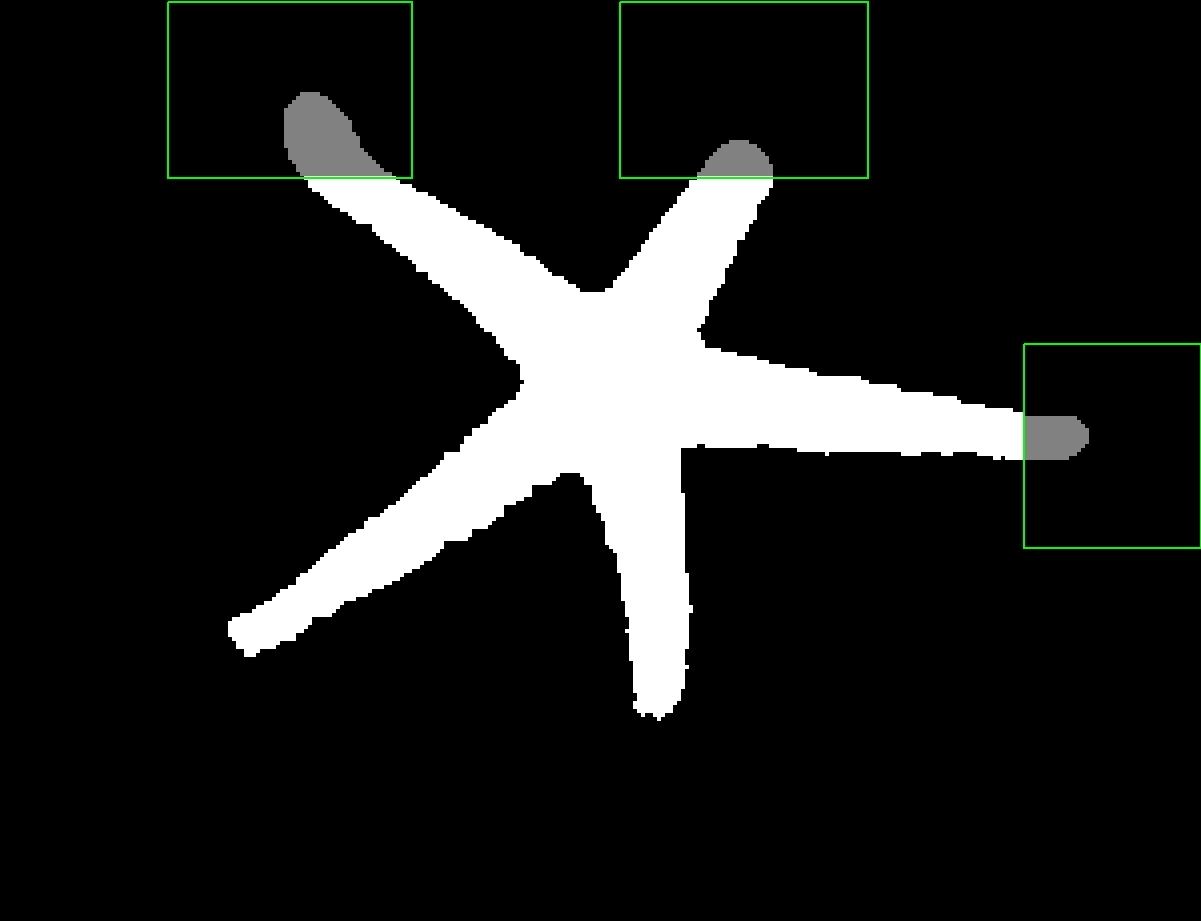}}&
\fbox{\includegraphics[width=\figwidth]{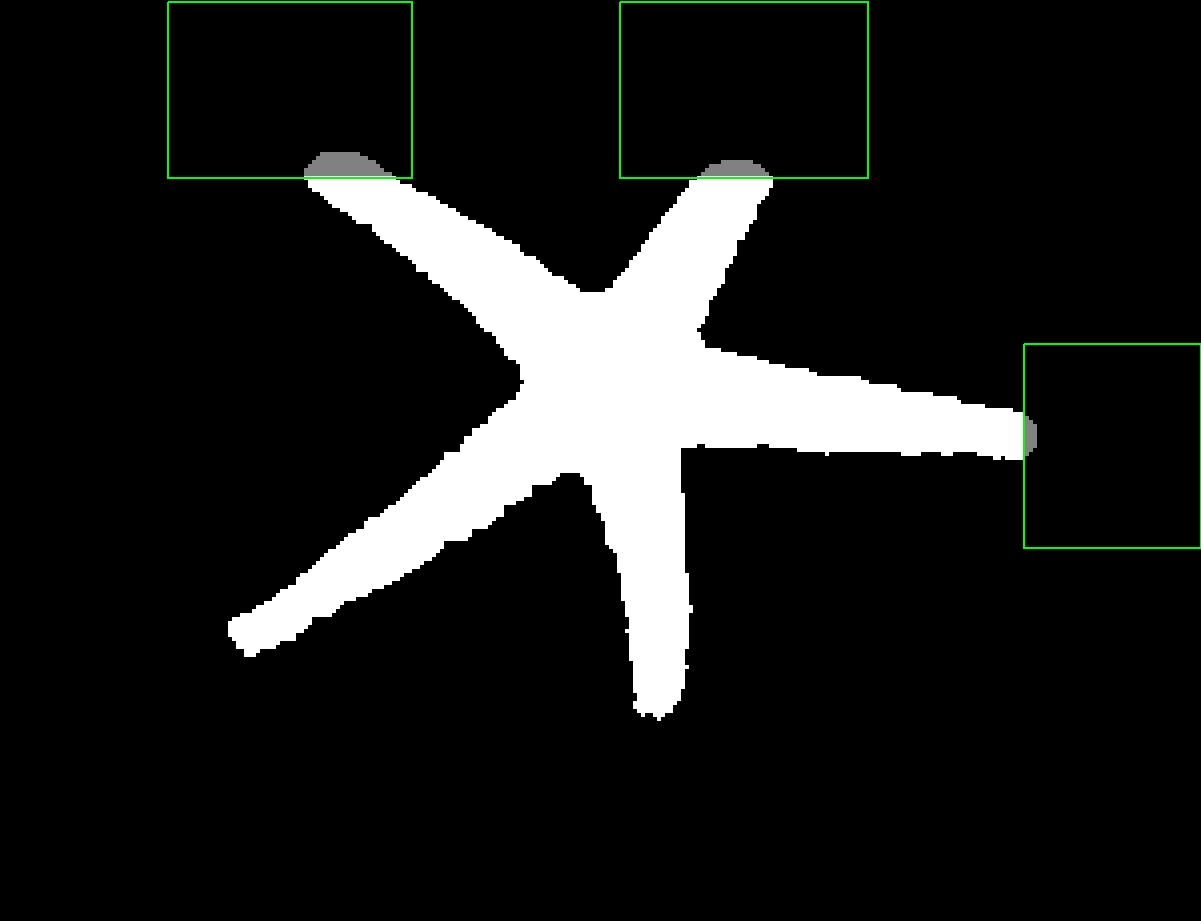}}\\
(a) & (b) & (c)
\end{tabular}
%
%\includegraphics[width = \linewidth, height = \linewidth]{fig/dummy}
%}
\end{center}
\caption{Combining length and curvature for inpainting: (a) pure curvature, (b) curvature + length, (c) curvature + more length.}
\label{fig_star_length}
\end{figure}
%
% It is seen that found solution for lines often have a lower energy than the ground truth solution, while they may deviate from a straight line. This can only happen as a result of inaccurate model, not as a result of suboptimality of optimization.
%Fig.~\ref{fig_inpaint_circle_lp} shows a relaxed labeling computed for a restricted inpainting problem of one of the circles.
%\input{tex/fig_inpaint_circle_lp.tex}
%
%
\mypar{Image Segmentation}
We use a simple model for the task of interactive FG/BG image segmentation, similar to \cite{BJ01}. Based on the user brush strokes (fig.~\ref{fig_giraff_task}(a)) we compute likelihoods using a Gaussian mixture model (GMM) with 10 components.
The difference of the unaries $\theta_v(1)-\theta_v(0)$ correspond to the negative log-likelihood ratio of foreground and background.
Fig.\ref{fig_giraff_task}(e) shows results when using a simple pairwise MRF (8-connectivity), which puts a prior on the length of the boundary.
By varying the strength of the prior we achieve various results, however, none of the results is satisfying. Note, the length prior is, in contrast to
\cite{BJ01}, not gradient-sensitive since the legs of the giraffe do not have an edge with sharp contrast.
Results for our curvature model for various strengths of the prior are shown in~fig.\ref{fig_giraff_task}(f). Note, no additional length prior is added. We clearly see that the curvature prior is able to properly segment the legs of the giraffe, compared to the length prior. Increasing the strength of the prior above some limit (1000) has almost no effect on the smoothness of the solution, because each local $8{\times}8$ window is already maximally smooth according to the model.
Note, that our result, e.g. fig.\ref{fig_giraff_task}(f,100), is visually superior to ~\cite{Schoenemann09}, fig. \ref{fig_giraff_task}(c), despite the fact that we use a grid with much coarser resolution (see detailed discussion below).

\begin{figure}[tr]
\small
\centering
\setlength{\doublerulesep}{0pt}
\setlength{\tabcolsep}{0pt}
\begin{tabular}{cc}
\phantom{(e)}&%
\setlength{\figwidth}{0.21\linewidth}
\setlength{\tabcolsep}{0.5pt}
\setlength{\doublerulesep}{0pt}
\setlength{\fboxsep}{0.3pt}
%\input{}
%\fbox{
\begin{tabular}{cccc}
\setlength{\tabcolsep}{0pt}
\begin{tabular}{c}
\fbox{\includegraphics[width=\figwidth]{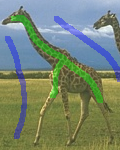}}
\end{tabular}&
\begin{tabular}{c}
\fbox{\includegraphics[width=\figwidth]{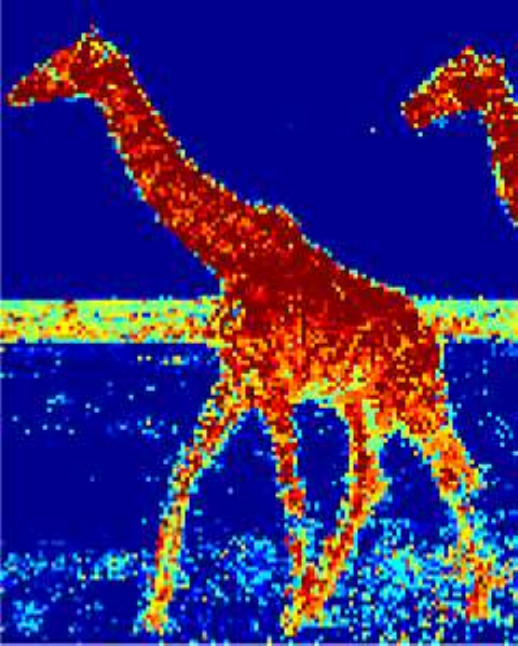}}
\end{tabular}&
\begin{tabular}{c}
\fbox{\includegraphics[width=\figwidth]{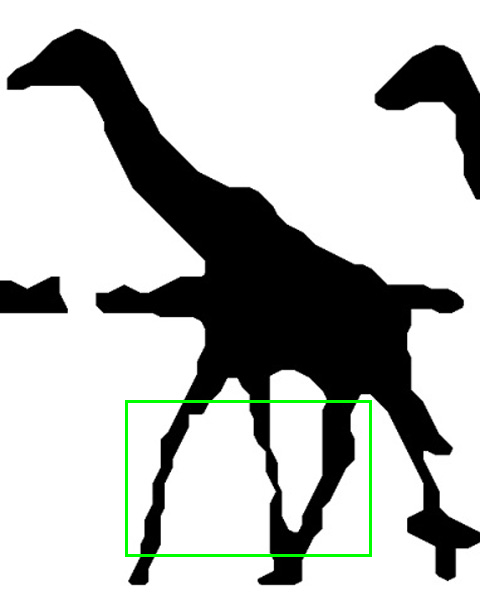}}
\end{tabular}&
\begin{tabular}{c}
\green \fbox{\includegraphics[width=\figwidth]{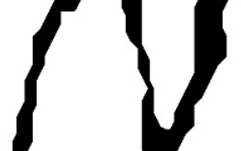}}\\
\red \fbox{\includegraphics[width=\figwidth]{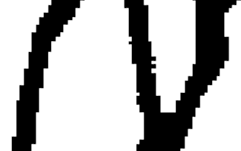}}
\end{tabular}\\
(a) & (b) & (c) & (d)
\end{tabular}
\end{tabular}
%
%%%%%%%%%%%%%%%
%
%
\setlength{\tabcolsep}{0pt}
\begin{tabular}{cc}
(e)&%
%\hspace{-5mm}
\setlength{\figwidth}{0.21\linewidth}
\setlength{\tabcolsep}{1pt}
\setlength{\doublerulesep}{0pt}
\setlength{\fboxsep}{0.3pt}
\begin{tabular}{ccccc}
\scriptsize
\fbox{\includegraphics[width=\figwidth]{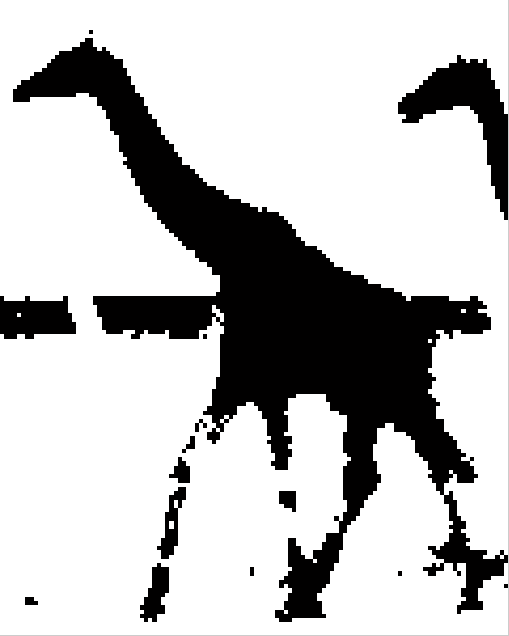}}&
%\fbox{\includegraphics[width=\figwidth]{fig/display/giraffe_002.png}}&
%\fbox{\includegraphics[width=\figwidth]{fig/display/giraffe_006.png}}&
\fbox{\includegraphics[width=\figwidth]{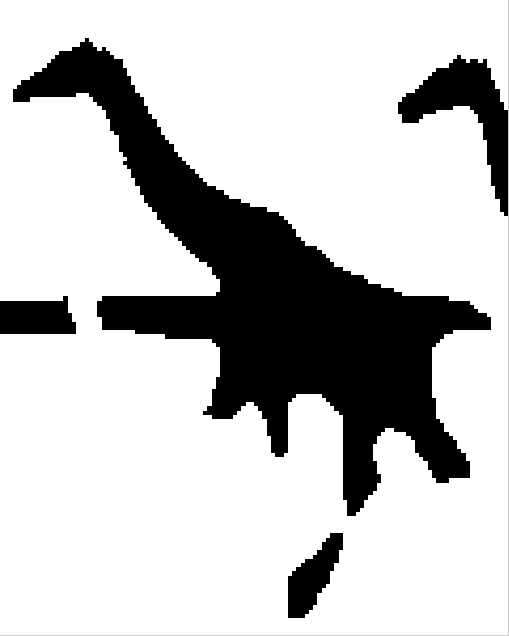}}&
\fbox{\includegraphics[width=\figwidth]{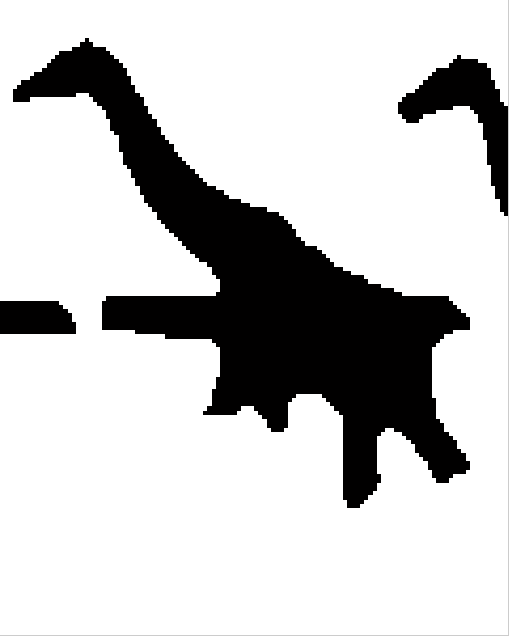}}&
\fbox{\includegraphics[width=\figwidth]{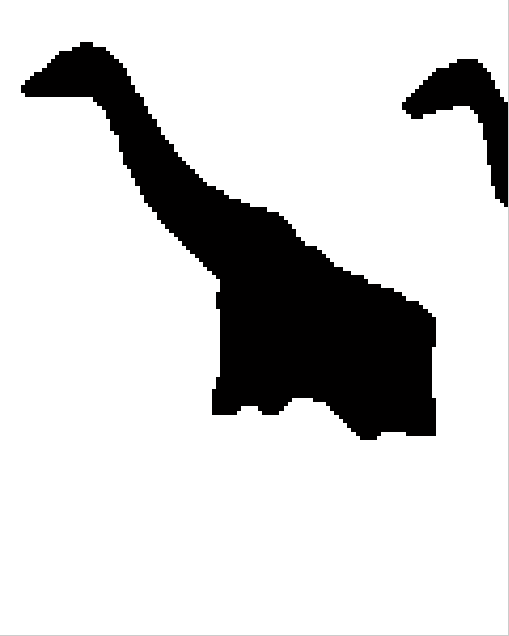}}\\
0.15 &   %0.3 &    %0.9&
1.0 &    1.5 &    3.0\\
\end{tabular}\\
%
%%%%%%%%%%%%%%%%%%%%%
%
(f)&%
%\hspace{-5mm}
\setlength{\figwidth}{0.21\linewidth}
\setlength{\tabcolsep}{1pt}
\setlength{\doublerulesep}{0pt}
\setlength{\fboxsep}{0.2pt}
\begin{tabular}{ccccc}
\scriptsize
\fbox{\includegraphics[width=\figwidth]{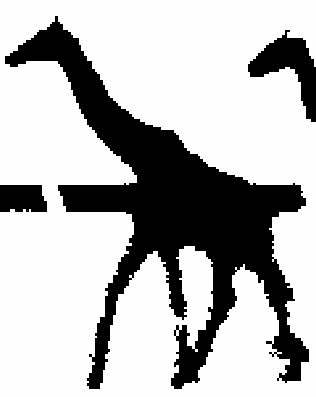}}&
\fbox{\includegraphics[width=\figwidth]{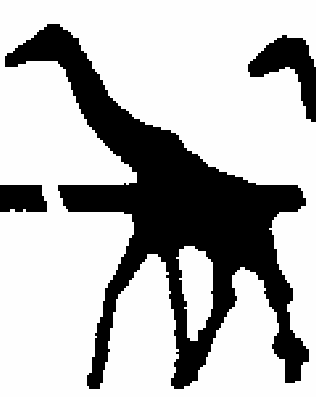}}&
%\fbox{\includegraphics[width=\figwidth]{fig/display/giraffe_model12_C3200_I400_icm_020}}&
\fbox{\includegraphics[width=\figwidth]{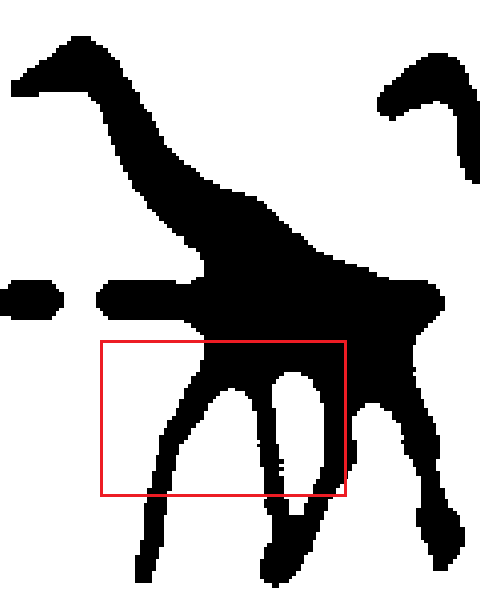}}&
%\fbox{\includegraphics[width=\figwidth]{fig/display/giraffe_model12_C3200_I400_icm_500}}&
\fbox{\includegraphics[width=\figwidth]{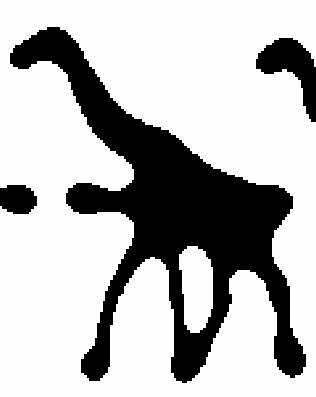}}\\
2 & 10 & %20 &
100 & %500 &
$>$1000\\
\end{tabular}
%%%%%%%%%%%%%%%%%%%%
\end{tabular}
\caption{\it {\bf Image segmentation.}
(a) Image with foreground (green) and background (blue) seeds;
(b) Color based unary potential costs (red implies foreground-favoring, blue implies background-favoring).
(c) Detail of segmentation result from~\cite{Schoenemann09} (top) and our result (f,100)
%Result copied from \cite{Schoenemann09}.
(d) Detail of segmentation~\cite{Schoenemann09} and our.
(e) Segmentation with length prior (8-connected model) for various strength of the prior (numbers below figure).
(f) Segmentation with curvature prior (our model) for various strength of the prior.
}
\label{fig_giraff_task}
\end{figure}
\begin{figure}[tr]
\begin{center}
\setlength{\tabcolsep}{2pt}
\setlength{\doublerulesep}{0pt}
\setlength{\fboxsep}{0.3pt}
%\input{}
%\fbox{
\setlength{\tabcolsep}{0pt}
\begin{tabular}[b]{cc}
\begin{tabular}[b]{c}\includegraphics[width=0.5\linewidth]{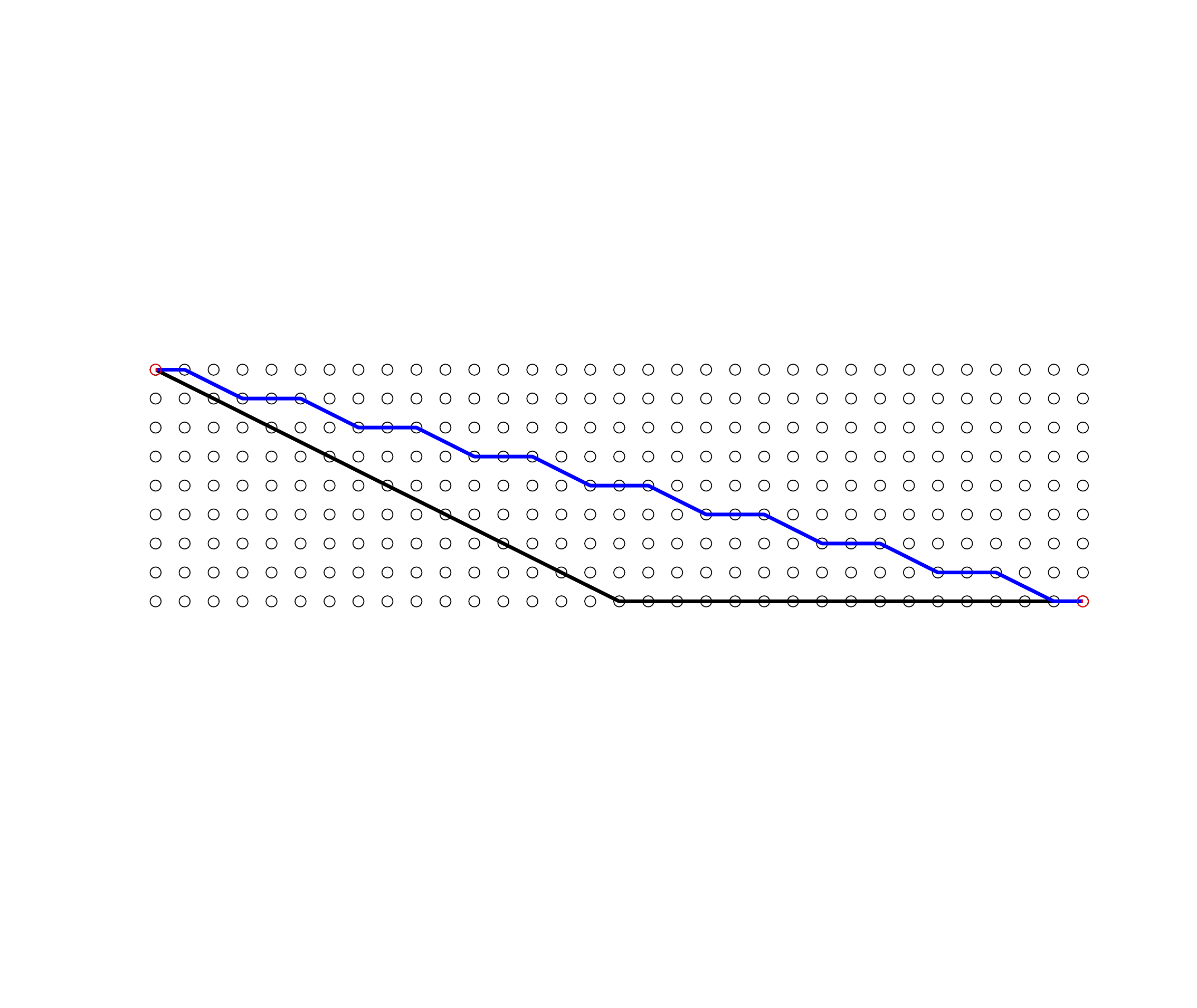}\\
(a)\\
\fbox{\includegraphics[width=0.4\linewidth]{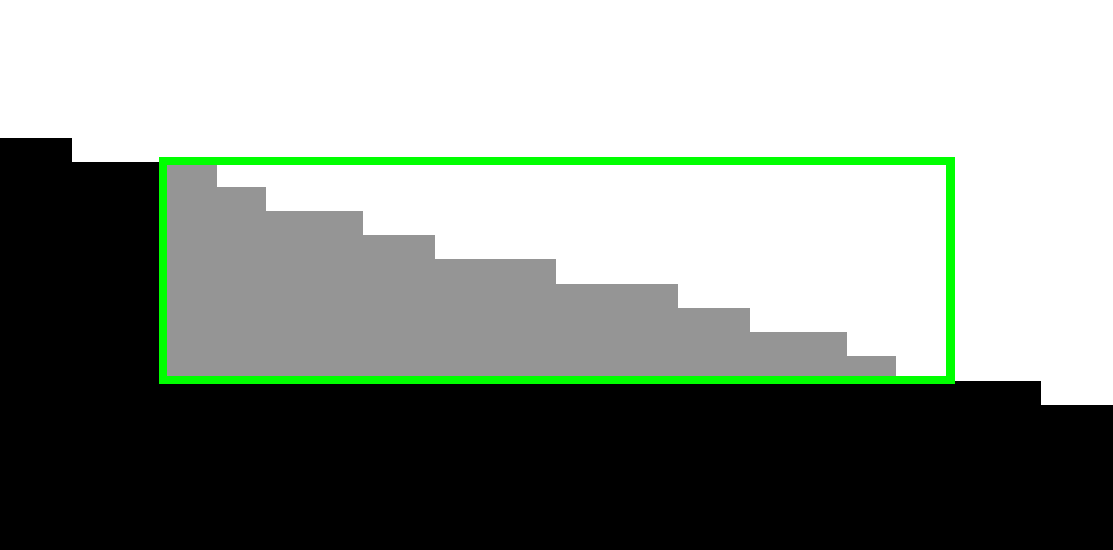}}\\
(c)\\
\end{tabular}\ &
\begin{tabular}[b]{c}\includegraphics[width=0.4\linewidth]{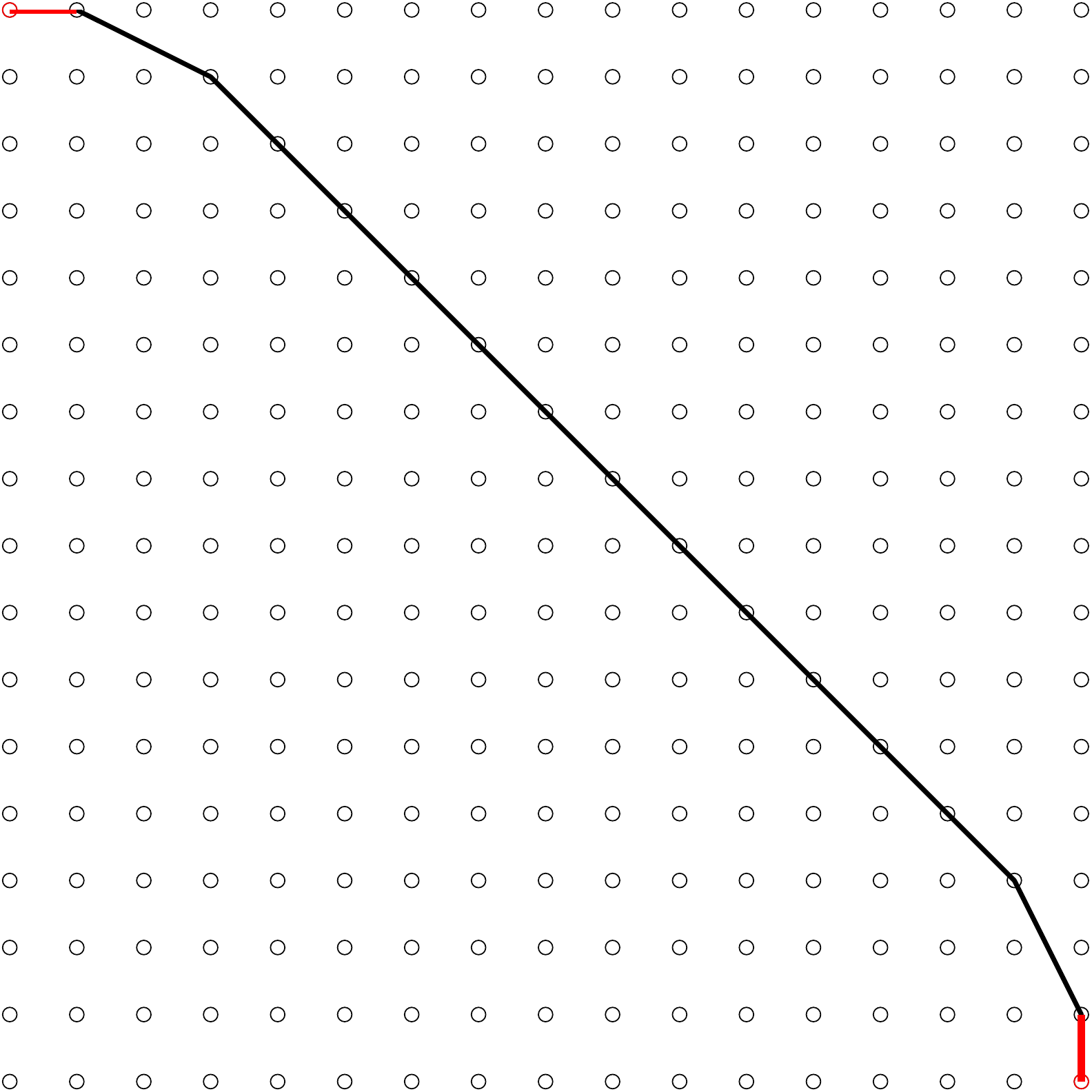}\\
(b)\\
\end{tabular}\\
\end{tabular}
\end{center}
\caption{{\bf Properties of \cite{Schoenemann09}}. In (a,b) the problem is to find the optimal shape with lowest curvature given the boundary conditions: the 2 end-points in (a) and 2 terminal edge-elements in (b). In (a,b) 
the black line shows an optimal solution of model~\cite{Schoenemann09}. The blue line in (a) is the closest approximation of a straight line. 
Note, in both cases (a,b) the model~\cite{Schoenemann09} has multiple solutions: (a) two solution with each having one corner point; (b) a family of optimal solutions with each having 4 corner points. (c) Inpaiting results by our model for the same problem as in (a). 
}
\label{fig_schoenemann_inpaint}
\end{figure}

\begin{figure}[tr]
\begin{center}
\setlength{\tabcolsep}{2pt}
\setlength{\doublerulesep}{2pt}
\setlength{\fboxsep}{0.3pt}
\begin{tabular}{cc}
\fbox{\includegraphics[width=0.45\linewidth]{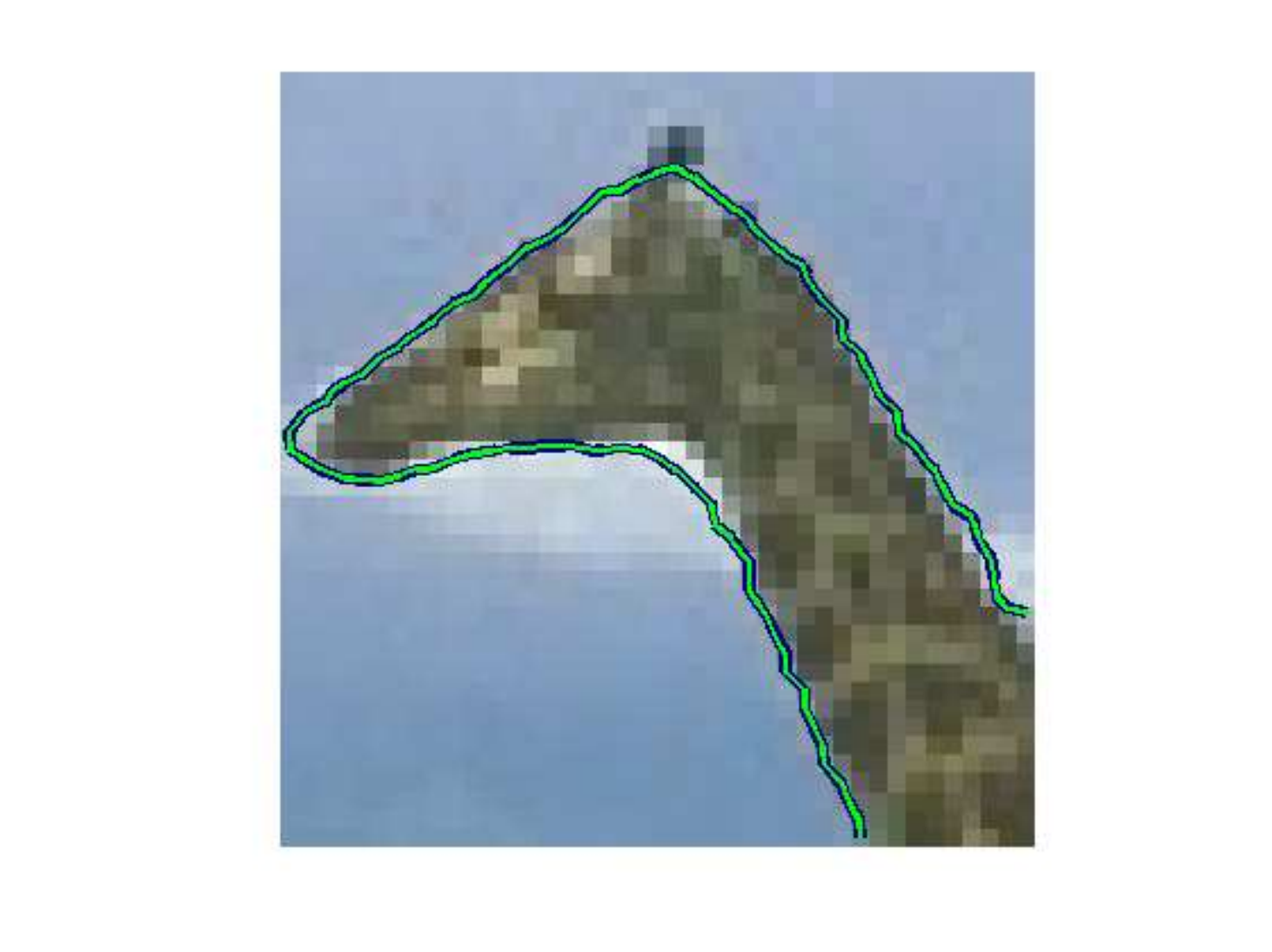}}&
\fbox{\includegraphics[width=0.45\linewidth]{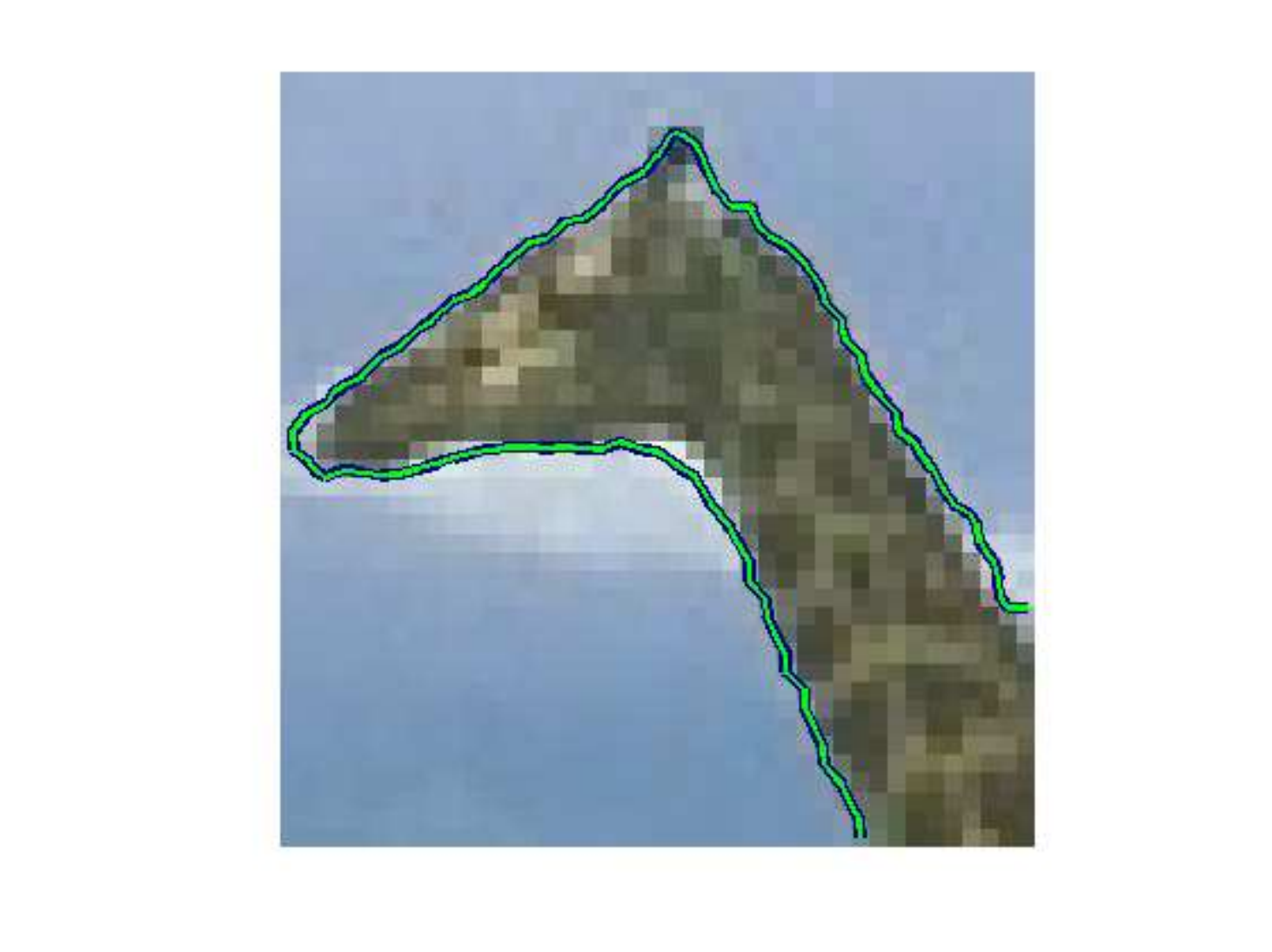}}\\
(a) & (b)
\end{tabular}
\end{center}
\caption{(a) Segmentation of the giraffe's head, where we show the zero level line of tree min-marginals. The curvature model smoothes out the giraff's ear. (b) Segmentation result with one pattern added (shown in the corner).}
\label{fig_ear}
\end{figure}

 %
\mypar{Analysis of the model of Sch\"onemann et al. \cite{Schoenemann09}}
The main property of the model of~\cite{Schoenemann09} is that there is a pre-defined set of quantized directions. For our analysis, we considered a restricted scenario (Fig.~\ref{fig_schoenemann_inpaint}) where it is evident that the optimal shape has to be described by a path in the graph. We used a 16-connected graph. For two consequent edge elements of the boundary, we approximated the squared curvature as $A^2\frac{l_1+l_2}{l_1 l_2}$, \cite{Bruckstrein07} (functional $G_2$), where $A$ is the angle between the line segments and $l_1, l_2$ their lengths. This is similar to~\cite{Schoenemann09} but a symmetric form. Another (not essential) difference of our re-implemtation of~\cite{Schoenemann09} is that they construct edge elements by subdividing each pixel, whereas we model edge element by end-points in a discrete grid (where edges can also intersect).
Fig.~\ref{fig_schoenemann_inpaint} reveals the problem of discritized directions. We observed that lines at directions which are not perfectly modeled in ~\cite{Schoenemann09} (e.g. the line at $1/4$ slope in fig.~\ref{fig_schoenemann_inpaint}(a)) have a very large approximation error. Indeed, the best approximation to the line in fig.\ref{fig_schoenemann_inpaint}(a) has many small ``steps'', whereas the optimal boundary in the model of~\cite{Schoenemann09} makes only one large ``step''. We believe that this effect is the reason for the visual artifacts in the segmentation result of the giraffe in fig. \ref{fig_giraff_task}(c), where the legs are approximated with a few straight lines.
\mypar{Generic Patterns}
As a demonstration of the extendibility of our model we made the following simple experiment. The giraff's ear is smoothed out by our model (fig.~\ref{fig_ear}(a)), since it is of high curvature and has weak support in the color model. We included one additional pattern which fits well to the ear. As all other patterns, this new pattern is available in all locations. The segmentation of the head, fig.~\ref{fig_ear}(b), clearly improves around the ear.
\section{Conclusions and Discussion}\label{discussion}
%\sectionskip
This paper shows how to compute compact representations of higher order priors which enable the use of standard algorithms for MAP inference. We have demonstrated our method on the problem of learning a `curvature-based' shape prior for image inpainting and segmentation. Our higher-order shape prior operates on a large set (neighbourhhod) of pixels and is less sensitive to discretization artifacts.
%The applicability of our method is not limited to image segmentation and inpainting; it can be used to obtain tractable representations for higher order priors for general labelling problems such as optical flow, stereo, and image restoration.
The applicability of our method is not limited to 2D image segmentation and inpainting; it could also be used for 3D completion. More generally, it can be used to obtain tractable representations for higher order priors for general labelling problems such as optical flow, stereo, and image restoration.

It would be interesting to extend our approach to incorporate other types of local shape properties, not necessarily defined by an analytic function but perhaps by exemplars. Such a generalization would very likely require a more general learning technique, which is an interesting direction for future research.

\appendix
\section{Solvable subclass}\label{subclass}
Here we give a short review of the result in~\cite{Kohli-Ladicky-Torr-09}. We show that a higher-order term of binary variables which is a composition of a piecewise-linear concave $\Real\to\Real$ function and a modular function with non-negative coefficients can be represented as min-projections of submodular quadratic functions.
In the context of model~\eqref{energy}, each higher-order term~\eqref{HOterm} must be of the form
%In a special case when each higher-order term of~\eqref{energy} $E_h$ can be represented as
\begin{equation}
E_h(x) = \min_{y}(a_y\<l,\x_{V(h)}\>+c_y),
\end{equation}
where $l\in \Real^{K^2}_+$ and $a$ and $c$ are arbitrary. In this case problem~\eqref{problem1}, where pairwise terms are submodular as well, is solvable in polynomial time.
Note, obviously one of the key differences to our general model is that the weight vector $\w_y$ in~\eqref{HOterm} can have arbitrary entries for different $y$, and of different signs for a single $y$. This is a crucial component for modeling some aspects, e.g. curvature, of the boundary of a segmentation.
\par
Consider the problem
\begin{equation}\label{problem1}
\min_{x\in {\{0,1\}}^\V}\big[ A(x)+P(x)+G(x)\big],
\end{equation}
where $A(x)$ is a linear term: $A(x) = \sum_{s\in\V}a_s x_s$ and $P(x)$ is quadratic and submodular. We are seeking for functions $G$ which can be represented as
\begin{equation}
\min_{y\in {\{0,1\}}^{\V{_{\rm y}}}}\big[ B(x)+C(y)+Q(x,y)+D\big],
\end{equation}
where $B$ and $C$ are linear and $Q$ is quadratic submodular. In this case problem~\eqref{problem1} would reduce to minimization of a submodular quadratic function, which is easily solvable by max-flow/min-cut.
\par
Functions suggested by~\cite{Kohli-Ladicky-Torr-09} may be described as follows. A function of the form
\begin{equation}
G(x) = \min(0,L(x)+D),
\end{equation}
where $L(x)= \sum_{s} l_s x_s$ has all coefficients $l_s$ non-positive and $D\in\Real$ can be represented as
\begin{equation}
\begin{split}
G(x) = \min_{y_1\in{\{0,1\}}} y_1(L(x)+D) \\
= \min_{y_1\in{\{0,1\}}}\Big[\sum_{s}l_s x_s y_1+Dy_1\Big],
\end{split}
\end{equation}
where the quadratic term in $(x,y)$ is submodular. In the case when all coefficients $l_s$ are non-negative, it can be represented as
\begin{equation}
\begin{split}
\min_{y_1\in{\{0,1\}}} (1-y_1)(L(x)+D) \\
= \min_{y_1\in{\{0,1\}}}L(x)+D+\Big[\sum_{s}-l_s x_s y_1-Dy_1\Big],
\end{split}
\end{equation}
 where the quadratic term is again submodular. This allows us to represent also function $\min(L^1(x)+D^1,L^2(x)+D^2) = \min(L^1(x)-L^2(x)+D^1-D^2,0)+L^2(x)+D^2$, under the condition that coefficients of $L^1-L^2$ are all non-positive or all non-negative.
 \par\
 \par
When coefficients of $L^1-L^2$ are of indefinite sign it seems impossible to represent $\min(L^1,L^2)$. To give an example, why it is so restrictive, consider
$G(x_1,x_2) = \min(x_1,x_2)$ This function is not submodular, indeed, $G(1,1)+G(0,0)=1 \ngeq 0 = G(1,0)+G(0,1)$ and therefore can not be represented as min-projection of a submodular one.
\par\
\par
Now, consider a more limited case, when $L^1 = a^1\sum_{s}l_s x_s+b^1$ and $L^2 = a^2\sum_{s}l_s x_s+b^2$, where $l_s\geq 0$. Then $\min(L^1,L^2)$ is always representable, because coefficients $L^1-L^2$ are all either positive or negative, depending on the sign of $a^1-a^2$. It is also easy to see that any $\Real\to \Real$ concave piece-wise linear function of $\sum_{s}l_s x_s$ can be represented as sum of minima where each minimum is of two linear functions satisfying the conditions and thus it is itself representable.
\section{Accounting for overlap}\label{overlap}
Let ${\rm bnd}(\x)$ be the set of boundary locations. It is easy to see that for a closed discrete contour, the number of points on the contour, $|{\rm bnd}(\x)|$ is the same as the number of edges in a 4-connected grid (i.e. the number of neighboring pixels with different labels).
Clearly, if $E_h(\x)$ is approximating the cost of the curvature in the neighborhood of locations $h$, then the sum~\eqref{bndsum} will be an inaccurate approximation of the continuous integral, at least because it measures the length in the 4-connected graph metric.
%
%\noindent {\bf Learning procedure for the full MRF.}
However, because each $8 \times 8$ pattern actually ``sees'' a larger neighborhood of the boundary, its weights may be adjusted so that the sum of pattern costs approximate the desired integral better. For example, weights of the pattern matching a diagonal line at 45 degrees may be scaled by $\frac{1}{\sqrt{2}}$. Because neighboring patterns overlap and because they have to model any arbitrary complex boundary, such an adjustment has to be done jointly for all patterns.
%Here the problem More generally, 
%This is one aspect, and another aspect is that the $8 \times 8$ patterns do overlap, hence an adjustment of the pattern weights could potentially improve on the overall accuracy of the model.
Note, the main focus of~\cite{Roth-09-FoE} was to address learning of overlapping terms, in the context of field of experts. 

We now propose a second training method, called Algorithm 2, which attempts to deal with these problems. The goal is to adjust the pattern weights such that the total curvature cost of a shape is approximated as well as possible.
\par
In particular, given a set of larger images, indexed by $i$, e.g. of size $100 \times 100$, where each image depicts a different continuous shapes $S^i$, with discretization $\x^i$, and total curvature $t^i$, we now formulate the learning problem as:
\begin{equation}\label{total_approx}
\arg \min_{\w,\bc} \sum_{i} \Big|\sum_h E_h(\x^i) - t^i \Big|.
\end{equation}
Let us use the same trick as in Alg.~1. If we linearize all terms $E_h$ by fixing the current best pattern for each location in each image, the problem simplifies. We therefore can iterate the following two steps:
\begin{minipage}[c]{\linewidth}
\parbox{\linewidth}{\center \bf Algorithm 2}
\noindent{\bf Input}: $\x^i$, $t^i$, $\w,\bc$\\
{\bf 1.} For all training images $i$ find matching patterns at all locations $h \in \U$:
\begin{equation}\label{find_all_patterns}
y^i_h = \arg\min\limits_{y}[ \sum_{v\in V(h)} w_{y,v} x_v^i + c_y]
\end{equation}
{\bf 2.} Refit all patterns:\\
\begin{equation}\label{total_refit_step}
\begin{array}{ll}
& (\w, \bc) = \\
& \arg\min\limits_{\w,\bc} \sum\limits_{i} \Big| (\sum_h \sum_{v
\in V(h)}w_{y^i_h,v} x^i_v +c_{y^i_h}) - t^i \Big|\\
& \vphantom{H^{H^H}_{H_H}} \st [ \min\limits_{\x \in \{0,1\}^{K^2}} \w_j^T \x ] +c_j \geq 0\ \ \ \forall j\in P \ .
\end{array}
\end{equation}
%\noindent{\bf Output}: $w,c$
\end{minipage}
The constraint in the second step enforces that the pattern cost for each arbitrary labeling is at least $0$, which is the lowest value of the function $f$. The main difference to Alg.~1 is that the refitting step does no longer decouple into estimating patterns weights independently. Step~\eqref{total_refit_step} will now be solved with an LP.
%
%In practice (see sec.~\ref{experiments}), we only used this algorithm to retrain parameters $\bc$, which was sufficient to compensate for the length bias.

\par
\begin{figure}[tr]
\begin{center}
\setlength{\tabcolsep}{0pt}
\setlength{\doublerulesep}{0pt}
\includegraphics[width=0.6\linewidth]{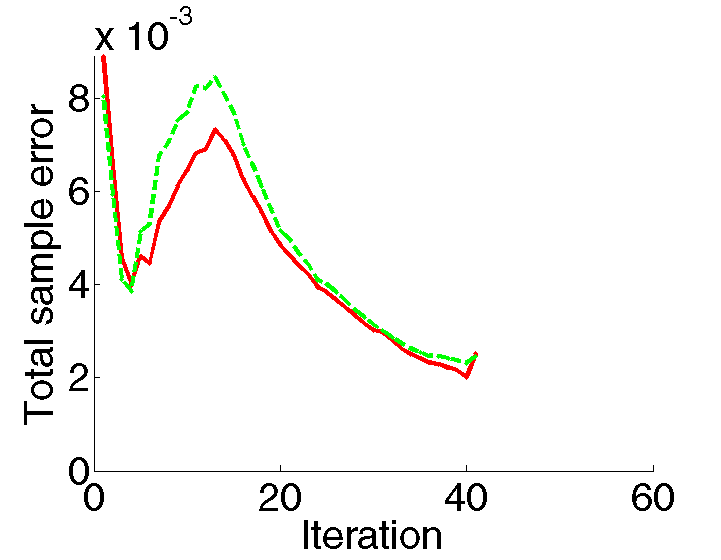}
\end{center}
\caption{Objective of~\eqref{total_approx} during iterations of Alg.~2. Red curve is evaluated on training data and green dashed curve is evaluated on the independent test data.
}
\label{fig:alg2_progress}
\end{figure}
\begin{figure}[tr]
\begin{center}
\setlength{\tabcolsep}{0pt}
\setlength{\doublerulesep}{0pt}
%\input{}
%\fbox{
\begin{tabular}{cc}
\includegraphics[width=0.5\linewidth]{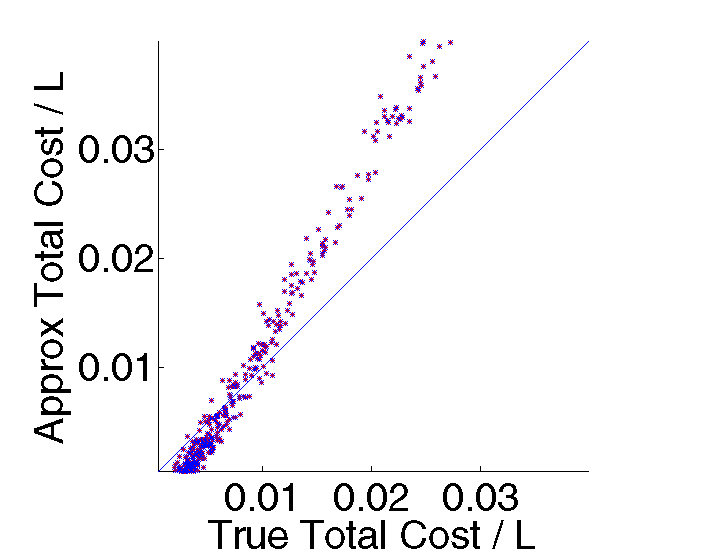}&
\includegraphics[width=0.5\linewidth]{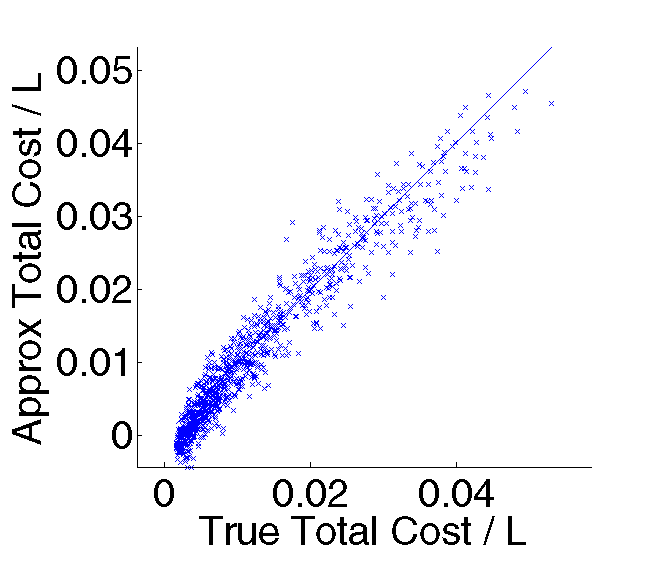}\\
(a) & (b)\\
\end{tabular}
%
%\includegraphics[width = \linewidth, height = \linewidth]{fig/dummy}
%}
\end{center}
\caption{Results after applying Alg.~2 to refit constant terms. We show approximate total cost / length vs. true total cost / length, for circles (a) and Fourier shapes (b) respectively. As can be seen teh bias is removed}
\label{fig_total_cost_refit}
\end{figure}

%Figure~\ref{fig_total_cost}(a,b) shows a bias of our model to overestimate the true cost. One of the reasons behind this is the fact that Manhattan length is larger than Euclidean. While this is not essential for shape optimization (makes a slightly different prior), we show that the model actually can compensate to some extent for this bias.
%
We tested Alg.~2 to retrain costs $\bc$ on the class of Fourier shapes (see fig \ref{fig_pointwise}g(right)) so as the total cost assigned by the model would match better to the true cost. It is initialized with patterns $(\w, \bc)$ which were learned using Alg.~1. Here we kept the weights $\w$ fixed and only update the weights $\bc$ in step 2. Figure~\ref{fig:alg2_progress} shows the progress of Alg.~2. Figure~\ref{fig_total_cost_refit} evaluats the new trained model on independent samples of circles and Fourier shapes. We see that the bias of the model to overestimate the total cost was reduced, compared to fig. \ref{fig_pointwise}(d,e). As expected, this is especially true for the Fourier shapes, since only Fourier shapes were used in training. Note, since retraining with Alg 2 gave only a mild improvement in approximation error, we did not use these patterns, but rather the ones obtained after Alg. 1 for all of our experiments.
\nocite{
Bertozzi07,
Chan01-CCD,
Kohli-Ladicky-Torr-09,
Komodakis09,
Rother09,
Schoenemann09,
Kolmogorov-06-convergent-pami,
Wainwright03nips,
Bruckstrein07,
Schoenemann-Cremers-07b,
Werner-PAMI07,
BJ01,
%Klodt08,
Esedoglu03-cont-segment-curv}

{\footnotesize
\bibliographystyle{splncs}
\bibliography{bib/references}
}

\end{document}